\documentclass{article}

\usepackage{subcaption}
\usepackage[accepted]{icml2024}

\usepackage[T1]{fontenc}
\usepackage[latin9]{inputenc}
\usepackage{amsmath}
\usepackage{amsthm}
\usepackage{amssymb}
\usepackage{xcolor}
\usepackage{bm}
\usepackage{wrapfig}
\usepackage{tcolorbox}
\usepackage{booktabs}

\usepackage{graphicx}

\usepackage[colorlinks=true, linkcolor=niceblue]{hyperref}
\usepackage[capitalize,noabbrev]{cleveref}

\definecolor{niceblue}{rgb}{0.10, 0.14, 0.76} 

\definecolor{nicered}{rgb}{0.70, 0.0, 0.0} 

\AtBeginDocument{%
\hypersetup{
    citecolor=niceblue,
    linkcolor=red,   
    urlcolor=niceblue,
    linktoc=none}
}

\theoremstyle{plain}
\newtheorem{thm}{\protect\theoremname}
\theoremstyle{plain}
\newtheorem{defn}{\protect\definitionname}
\theoremstyle{plain}

\theoremstyle{plain}

\theoremstyle{plain}
\newtheorem{prop}{Proposition}
\theoremstyle{plain}

\theoremstyle{plain}
\newtheorem{assumption}{Assumption}

\newcommand{\init}[1]{\texttt{Init[#1]}}

\makeatother

\usepackage{babel}
\providecommand{\definitionname}{Definition}
\providecommand{\examplename}{Example}
\providecommand{\theoremname}{Theorem}

\usepackage[textsize=tiny]{todonotes}

\icmltitlerunning{Efficient Low Rank Adaptation}

\begin{document}
\global\long\def\reals{\mathbb{R}}%
\global\long\def\data{\mathcal{D}}%
\global\long\def\normal{\mathcal{N}}%
\global\long\def\E{\mathbb{E}}%
\global\long\def\loss{\mathcal{L}}%
\global\long\def\bigO{\mathcal{O}}

\twocolumn[
\icmltitle{LoRA+: Efficient Low Rank Adaptation of Large Models}



\icmlsetsymbol{equal}{*}

\begin{icmlauthorlist}
\icmlauthor{Soufiane Hayou}{equal,yyy}
\icmlauthor{Nikhil Ghosh}{equal,comp}
\icmlauthor{Bin Yu}{comp}
\end{icmlauthorlist}

\icmlaffiliation{yyy}{Simons Institute, UC Berkeley}
\icmlaffiliation{comp}{Department of Statistics, UC Berkeley}


\icmlcorrespondingauthor{Soufiane Hayou}{hayou@berkeley.edu}
\icmlcorrespondingauthor{Nikhil Ghosh}{nikhil$\_$ghosh@berkeley.edu}

\icmlkeywords{Machine Learning, ICML}

\vskip 0.3in
]



\printAffiliationsAndNotice{\icmlEqualContribution} 

\begin{abstract}
In this paper, we show that Low Rank Adaptation (LoRA) as originally introduced in \cite{hu2021lora} leads to suboptimal finetuning of models with large width (embedding dimension). This is due to the fact that adapter matrices $A$ and $B$ in LoRA are updated with the same learning rate. Using scaling arguments for large width networks, we demonstrate that using the same learning rate for $A$ and $B$ does not allow efficient feature learning. We then show that this suboptimality of LoRA can be corrected simply by setting different learning rates for the LoRA adapter matrices $A$ and $B$ with a well-chosen fixed ratio. We call this proposed algorithm LoRA$+$. In our extensive experiments, LoRA$+$ improves performance ($1\%-2\%$ improvements) and finetuning speed (up to $\sim2$X SpeedUp), at the same computational cost as LoRA.
\end{abstract}

\section{Introduction}

State-of-the-art (SOTA) deep learning models all share a common characteristic: they all have an extremely large number of parameters (10's if not 100's of billions parameters). Currently, only a few industry labs can pretrain large language models due to their high training cost. However, many pretrained models are accessible either through an API (GPT4, \cite{openai2023gpt4}) or through open-source platforms (Llama, \cite{touvron2023llama}). Most practitioners are interested in using such models for specific tasks and want to \emph{adapt} these models to a new, generally smaller task. This procedure is known as \emph{finetuning}, where one adjusts the weights of the pretrained model to improve performance on the new task. However, due to the size of SOTA models, adapting to down-stream tasks with full finetuning (finetuning all model parameters) is computationally infeasible as it requires modifying the weights of the pretrained models using gradient methods which is a costly process.  Besides, a model that has already learned generally useful representations during pretraining would not require  in-principle significant adaptation of all parameters. With this intuition, researchers have proposed a variety of resource-efficient finetuning methods which typically freeze the pretrained weights and tune only a small set of newly inserted parameters. Such methods include prompt tuning \citep{lester2021power} where a ``soft prompt" is learned and appended to the input, the adapters method \citep{houlsby2019parameter} where  lightweight ``adapter" layers are inserted and trained, and $(IA)^3$ \citep{liu2022few} where activation vectors are modified with learned scalings. Another resource-efficient method is known as \emph{Low Rank Adaptation} \citep{hu2021lora}, or simply LoRA. In LoRA finetuning, only a low rank matrix, called an \textit{adapter}, that is added to the pretrained weights is trainable. The training can be done with any optimizer and in practice a common choice is Adam \citep{kingma2014adam}. Since the trained adapter is low-rank, this effectively reduces the number of trainable parameters in the fine-tuning process, significantly decreasing the training cost. On many tasks such as instruction finetuning, LoRA has been shown to achieve comparable or better performance compared with full-finetuning \citep{wang2023far, liu2023improved}, although on complicated, long form generation tasks, it is not always as performant. The impressive performance and the computational savings of LoRA have contributed to it becoming an industry standard finetuning method. 

Efficient use of LoRA requires a careful choice of hyperparameters: the rank and the learning rate. While some theoretical guidelines on the choice of the rank in LoRA exist in the literature (see e.g. \citet{zeng2023expressive}), there are no principled guidelines on how to set the learning rate, apart from common choices  of order $1e$-$4$.
\begin{figure}
    \centering
    \includegraphics[width=.9\linewidth]{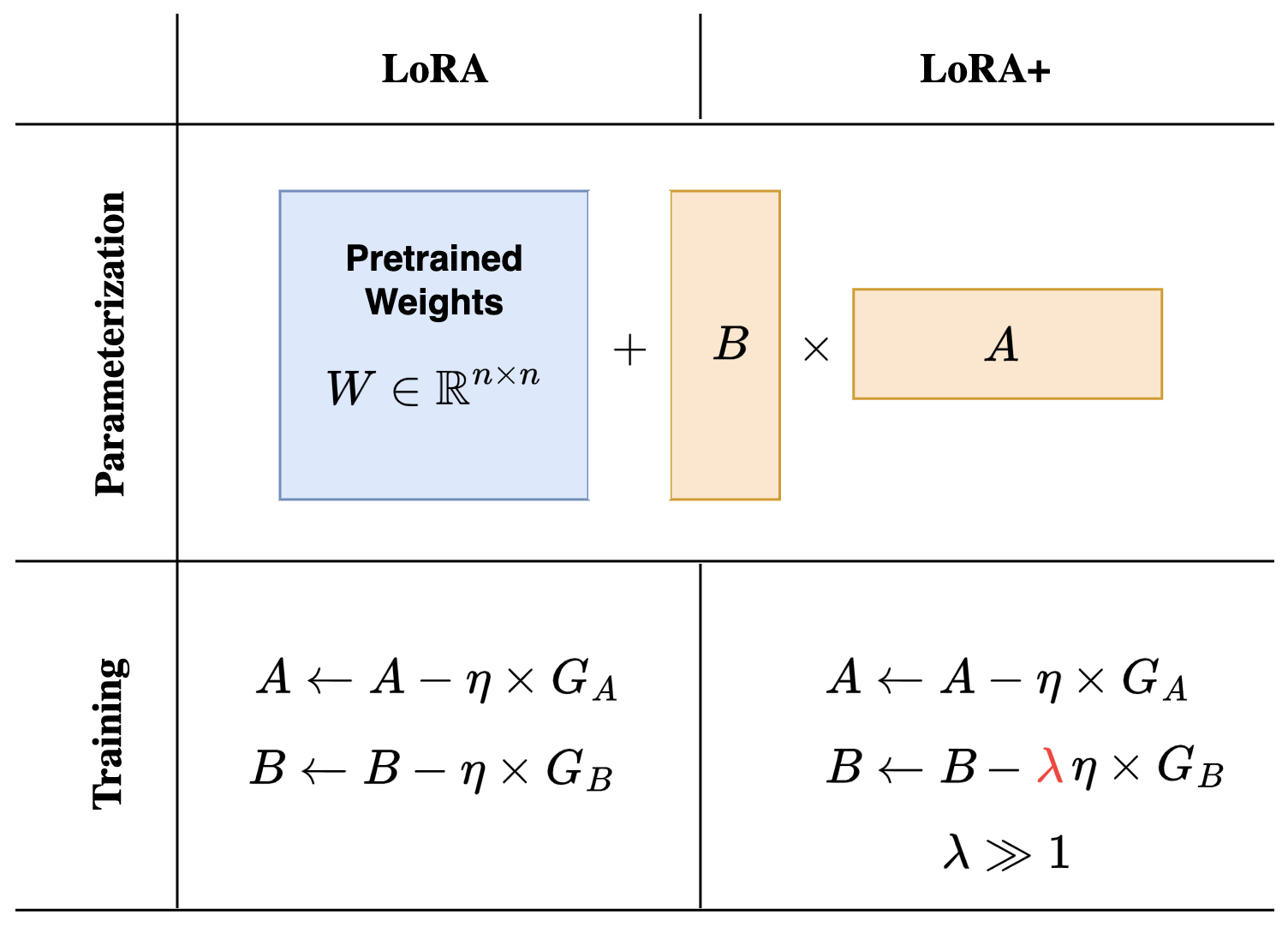}
    \caption{\small{The key difference between standard LoRA and LoRA$+$ is in how learning rates are set (the matrices $G_A$ and $G_B$ are `effective' gradients from AdamW) With standard LoRA, the learning rate is the same for $A$ and $B$, which provably leads to suboptimal learning when embedding dimension is large. In LoRA$+$, we set the learning rate of $B$ to be $\lambda \times$ that of $A$, where $\lambda \gg 1$ is fixed. We later provide guidelines on how to set $\lambda$.}}
    \vspace{-0.5cm}
    \label{fig:intro_table}
\end{figure}
\vspace{-0.3cm}
\paragraph{Related Work.} \citet{dettmers2023qlora} introduced a quantized version of LoRA (or QLoRA), which further reduces computation costs by quantizing pretrained weights down to as few as four bits. Using QLoRA enables fine-tuning Llama-65b \citep{touvron2023llama}, on a single consumer GPU while achieving competitive performance with full-finetuning. To further improve LoRA training with quantization, \citet{li2023loftq} introduced a new method called LoftQ for computing a better initialization for quantized training. Additional variations of LoRA have been proposed such as VeRA \citep{kopiczko2023vera} which freezes random weight tied adapters and learns vector scalings of the internal adapter activations. This achieves a further reduction in the number of trainable parameters while achieving comparable performance to LoRA on several NLP finetuning tasks. However, to the best of our knowledge, there is no principled guidance for setting LoRA learning rate which is the focus of our work. 
\vspace{-0.5cm}
\paragraph{Contributions.} We provide guidelines for setting the learning rate through a theory of scaling for neural networks. There is a significant number of works on the scaling of neural networks from the infinite width/depth perspective. The approach is simple: take the width/depth of a neural network to infinity,\footnote{Depending on the model, one might want to scale width with fixed depth and vice-versa, or both at the same time. See \cref{sec:scaling} for more details.} understand how the limit depends on the choice of the hyperparameters in the training process such as the learning rate and initialization variance, then derive principled choices for these hyperparameters to achieve some desired goal (e.g.\ improve feature learning). Examples of the infinite-width limit include works on initialization schemes such as \cite{he2016deep,yang2019scaling}, or more holistically network parametrizations such as \citep{yang2021tensor} where the authors introduced $\mu$P, a neural network parameterization ensuring feature learning in the infinite-width limit, offering precise scaling rules for architecture and learning rates to maximize feature learning. Examples for the depth limit include initialization strategies \citep{schoenholz2017deep, he2023deep, hayou19activation}, block scaling (see e.g. \cite{pmlr-v130-hayou21a,hayou2023on, noci2023shaped}), depth parametrizations \citep{yang2023depth,bordelon2023depthwise} etc. Here we propose to use the same strategy to derive scaling rules for the learning rate in LoRA for finetuning. More precisely, we study the infinite-width limit of LoRA finetuning dynamics and show that standard LoRA setup is suboptimal. We correct this by introducing a new method called LoRA$+$ that improves feature learning in low rank adaptation in the this limit. The key innovation in LoRA$+$ is setting different learning rates for $A$ and $B$ modules (LoRA modules) as explained in \cref{fig:intro_table}. Our theory is validated with extensive empirical results with different language of models and tasks.

\section{Setup and Definitions}
Our methodology in this paper is model agnostic and applies to general neural network models. Let us consider a neural network of the form
\begin{equation}\label{eq:model}
\begin{cases}
Y_{in}(x) = W_{in}x,\\
Y_l(x) = \mathcal{F}_l(W_l,Y_{l-1}(x)), \; l\in[L],\\
Y_{out}(x) = W_{out} Y_L(x), 
\end{cases}
\end{equation}
where $x\in\reals^{d}$ is the input, $L\geq1$ is the network depth,
$(\mathcal{F}_{l})_{l\in[L]}$ are mappings that define the layers, $W_{l}\in\reals^{n\times n}$ are the hidden weights, where $n$ is the
network $\emph{width}$, and $W_{in}, W_{out}$ are input and output embedding weights. 

Model \eqref{eq:model} is \emph{pretrained} on some dataset $\data$ to perform some specified task (e.g. next token prediction). 
Once the model is pretrained, one can finetune it to improve performance on some downstream task. To achieve this with relatively small devices (limited GPUs), resource-efficient finetuning methods like LoRA significantly reduce the computational cost by considering low rank weight matrices instead of full rank finetuning (or simply full finetuning). 
\begin{defn}[Low Rank Adapters (LoRA) from \cite{hu2021lora}]\label{def:lora}
For any weight matrix $W\in\reals^{n_{1}\times n_{2}}$ in the pretrained
model, we constrain its update in the fine-tuning process by representing
the latter with a low-rank decomposition $W=W^{*}+\frac{\alpha}{r}BA$.
Here, only the weight matrices $B\in\reals^{n_{1}\times r}$,
$A\in\reals^{r\times n_{2}}$ are trainable. The rank $r\ll\min(n_{1},n_{2})$ and $\alpha\in\reals$
are tunable constants. 
\end{defn}
\paragraph{Scaling of Neural Networks.}
It is well known that as the width $n$ grows, the network initialization scheme and the learning should be adapted to avoid numerical instabilities and ensure efficient learning. For instance, the variance of the initialization weights (in hidden layers) should scale $1/n$ to prevent arbitrarily large pre-activations as we increase model width $n$ (e.g. He init \cite{he2016deep}). To derive such scaling rules, a principled approach consist of analyzing statistical properties of key quantities in the model (e.g. pre-activations) as $n$ grows and then adjust the initialization, the learning rate, and the architecture itself to achieve desirable properties in the limit $n \to \infty$ \citep{hayou19activation, deepinfoprop2017, yang2019scaling, yang2023tensor}. This approach is used in this paper to study feature learning dynamics with LoRA in the infinite-width limit. This will allow us to derive scaling rules for the learning rates of LoRA modules. For more details about the theory of scaling of neural networks, see \cref{sec:scaling}.
\vspace{-1em}
\paragraph{Notation.} Hereafter, we use the following notation to describe the asymptotic behaviour as the width $n$ grows. Given sequences $c_n \in \reals$ and $d_n \in \reals^+$, we write $c_n = \bigO(d_n)$, resp. $c_n = \Omega(d_n)$, to refer to $c_n < \kappa d_n$, resp. $c_n > \kappa d_n$, for some constant $\kappa > 0$. We write $c_n = \Theta(d_n)$ if both $c_n = \bigO(d_n)$ and $c_n = \Omega(d_n)$ are satisfied. For vector sequences $c_n = (c_n^i)_{1 \leq i \leq k} \in \reals^k$ (for some $k >0$), we write $c_n = \bigO(d_n)$ when $c_n^i = \bigO(d_n^i)$ for all $i \in [k]$, and same holds for other asymptotic notations. Finally, when the sequence $c_n$ is a vector of random variables, convergence is understood to be convergence in second moment ($L_2$ norm).

\section{An Intuitive Analysis of LoRA}\label{sec:toy}
Our intuition is simple: the matrices $A$ and $B$ have ``transposed'' shapes and one would naturally ask whether the learning rate should be set differently for the two matrices. In practice, most SOTA models have large width (embedding dimension). Thus, it makes sense to study the training dynamics when the width goes to infinity. 
\subsection{LoRA with a Toy Model}
Consider the following linear model 
\begin{equation}\label{eq:toy_model}
f(x)=(W^*+b a^\top)x,
\end{equation}
where $W^* \in \reals^{1\times n}$ are the pretrained weights, $b\in\reals,a\in\reals^{n}$ are LoRA weights,\footnote{Here, we consider $n_2 = 1$ to simplify the analysis. All the conclusions remain essentially valid when $n_2 = n_1 = n$.} $x\in\reals^{n}$ is the model input. This setup corresponds to $n_1=1, n_2=n, r=1$ in \cref{def:lora}. We assume that the weights $W^*$ are fixed (from pretraining). The goal is to minimize the loss $\loss(\theta)=\frac{1}{2}(f(x)-y)^{2}$ where $\theta=(a,b)$ and $(x,y)$ is an input-output datapoint.\footnote{For simplicity, we assume that the finetuning dataset consists of a single sample. Our analysis is readily generalizable to multiple samples.} We assume that $x = \Theta_n(1)$ which means that input coordinates remain of the same order as we increase width. In the following, we analyze the behaviour of the finetuning dynamics as model width $n$ grows. 
\paragraph{Initialization.} We consider a Gaussian initialization of the weights as follows: $a_i \sim \normal(0,\sigma_a^2)$, $b \sim \normal(0, \sigma_b^2)$.\footnote{The Gaussian distribution can be replaced by any other distribution with finite variance.} With LoRA, we generally want to initialize the product $ba^\top$ to be $0$ so that finetuning starts from the pretrained model. This implies at least one of the weights $a$ and $b$ is initialized to $0$. If both are initialized to $0$, it is trivial that no learning occurs in this case since this is a saddle point. Thus, we should initialize one of the parameters $a$ and $b$ to be non-zero and the other to be zero. If we choose a non-zero initialization for $a$, then following standard initialization schemes (e.g., He Init \citep{he2016deep}, LeCun Init \citep{lecun2002efficient}), one should set $\sigma_a^2 = \Theta(n^{-1})$ to ensure $a^\top x$ does not explode with width. This is justified by the Central Limit Theorem (CLT).\footnote{Technically, the CLT only ensures the almost sure convergence, the $L_2$ convergence follows from the Dominated Convergence Theorem. We omit these technical details in this paper.} On the other hand, if we choose a non-zero initialization for $b$, one should make sure that $\sigma_b^2 = \Theta(1)$. This leaves us with two possible schemes:
\vspace{-0.2cm}
\begin{itemize}
    \item \init{1}: $\sigma_b^2 = 0, \sigma_a^2 = \Theta(n^{-1})$.
    \item \init{2}: $\sigma_b^2 = \Theta(1), \sigma_a^2 =0$.
\end{itemize}
\vspace{-0.2cm}
Our analysis will only consider these two initialization schemes for LoRA modules, although the results should in-principle hold for other schemes, providing that stability (as discussed above) is satisfied.
\paragraph{Learning rate.} WLOG, we can simplify the analysis by assuming that $W^* = 0$. This can be achieved by setting $\Tilde{y} = y - W^* x$. The gradients are given by
\begin{align*}
\frac{\partial\loss}{\partial b} =a^{\top}x(f(x)-y), \,\,\,
\frac{\partial\loss}{\partial a}  =b(f(x)-y)x.
\end{align*}
 We use subscript $t$ to denote the finetuning step. Let $U_t = (f_t(x)-y)$. At step $t$ with learning rate $\eta>0$, we have 
\begin{align*}
&\Delta f_t \overset{def}{=} f_t(x) - f_{t-1}(x) = -\underset{\delta_t^1}{\underbrace{\eta b_{t-1}^{2}U_{t-1}\|x\|^{2}}}\\
&-\underset{\delta_t^2}{\underbrace{\eta(a_{t-1}^{\top}x)^{2}U_{t-1}}}+\underset{\delta_t^3}{\underbrace{\eta^2 U_{t-1}^{2}b_{t-1}(a_{t-1}^{\top}x)\|x\|^{2}}}.
\end{align*}
The update in model output is driven by the three terms $(\delta_t^i)_{i \in \{1,2,3\}}$. The first two terms represent ``linear'' contributions to the update, i.e. change in model output driven by fixing $b$ and updating $a$ and vice-versa. These terms are order one in $\eta$. The third term $\delta_t^3$ represents a multiplicative update, compounding the updates in $a$ and $b$, and is an order two term in $\eta$. As $n$ grows, a desirable property is that $\Delta f_t=\Theta(1)$. Intuitively, this means that as we scale the width, feature updates do not `suffer' from this scaling (see \cref{sec:scaling} for more details). An example of a scenario where feature learning is affected by scaling is the lazy training regime \citep{jacot2018neural}, where feature updates are of order $\Theta(n^{-1/2})$ which implies that no feature learning occurs in the limit $n \to \infty$. The condition $\Delta f_t = \Theta(1)$ also implies that the update does not explode with width, which is also a desirable property. 

Having $\Delta f_t = \Theta(1)$ satisfied implies that at least one of the three terms $(\delta_t^i)_{i \in \{1,2,3\}}$ is $\Theta(1)$. Ideally, we want both $\delta^1_t$ and $\delta^2_t$ to be $\Theta(1)$ because otherwise it means that either $a$ or $b$ is not efficiently updated. For instance, if $\delta^1_t = o(1)$, it means that as $n \to \infty$, the model acts as if $a$ is fixed and only $b$ is trained. Similar conclusions hold when $\delta^2_t = o(1)$. Having both $\delta^1_t$ and $\delta^2_t$ being $\Theta(1)$ in width means that both $a$ and $b$ parameter updates significantly contribute to the change in $f_t(x)$, and we say that feature learning with LoRA is \emph{efficient} when this is the case, i.e. $\delta_i^t = \Theta(1)$ for $i \in \{1,2\}$ and all $t>1$. We will formalize this definition of efficiency in the next section. The reader might wonder why we do not require that $\delta^3_t$ be $\Theta(1)$. We will see that when both $\delta^1_t$ and $\delta^2_t$ are $\Theta(1)$, the term $\delta^3_t$ is also $\Theta(1)$.

\paragraph{Efficiency Analysis.} Let us assume that we train the model with gradient descent with learning rate $\eta = \Theta(n^c)$ for some $c \in \reals$, and suppose that we initialize the model with \init{1}. Sine the training dynamics are mainly matrix vector products, sum of vectors/scalars etc (see \citep{yang2022tensor}),\footnote{A crucial assumption for this to hold is also to have that for any matrix/vector product in the training dynamics, the product dimension (the dimension along which the matrix/vector product is calculated) is $\Theta(n^\alpha)$ for some $\alpha>0$. For instance, in the case of Transformers, this is satisfied since the MLP embedding dimension is generally $k \times n$. However, this condition would be violated if for instance one considers MLP embedding dimension $ k n \log(n)$. Such non-standard scaling choices require a particular treatment, but the conclusions remain the same.} it is easy to see that any quantity in the training dynamics should be of order $n^{\gamma}$ for some $\gamma \in \reals$. For any quantity $v$ in the training dynamics, we  write $v = \Theta(n^{\gamma[v]})$. When $v$ is a vector, we use the same notation when all entries of $v$ are $\Theta(n^{\gamma[v]})$. The $\gamma$ notation is formally defined in \cref{app:proofs}. 

Starting from initialization, we have $f_0(x) = 0$. LoRA finetuning is efficient when $\delta_t^1 = \Theta(1)$ and $\delta^2_t = \Theta(1)$ for all $t>1$,\footnote{Here we use the $t>1$ instead of $t>0$ because at $t\leq 1$, at least one the terms $\delta_1^1$ or $\delta_1^2$ will be zero.} and $f_t(x) = \Theta(1)$ for $t> 1$. This translate to 
\begin{equation*}
    \begin{cases}
        c+2\gamma[b_{t-1}] + 1 = 0 \quad (\delta^1_t = \Theta(1))\\
        c + 2 \gamma[a_{t-1}^\top x] = 0\quad (\delta^2_t = \Theta(1)) \\
        \gamma[b_{t-1}] + \gamma[a_{t-1}^\top x] = 0 \quad (f_{t-1}(x) = \Theta(1))\\        
    \end{cases}
\end{equation*}
Solving this equation yields $c = -1/2$, i.e. the learning rate should scale as $\eta = \Theta(n^{-1/2})$ in order to achieve efficient feature learning. At initialization, $b_0 = 0$ and $a_0^\top x = \Theta(1)$ (by Central Limit Theorem). Through an inductive argument, for $t >0$, $b_t$ will be of order $\Theta(n^{-1/2})$ and $a_t^\top x$ will be  of order $\Theta(1)$, yielding $f_t(x) = \Theta(n^{-1/2})$. Indeed, at each iteration the update to $b_t$ will be of order $\Theta(\eta y a_{t-1}^\top x) = \Theta(n^{-1/2})$ and the updates to $a_t$ are of order $\Theta(\eta b_{t-1} y x) = \Theta(n^{-1})$. As $f_t = \Theta(n^{-1/2})$, this yields a contradiction towards learning $\Theta(1)$ features.

This shows that we cannot have both $\delta_t^1$ and $\delta_t^2$ to be $\Theta(1)$ with this parametrization (also true with \init{2}). We formalize this result in the next proposition and refer the reader to \cref{app:proofs} for further technical details.

\begin{prop}[Inefficiency of LoRA fine-tuning]\label{prop:inefficiency_toy}
Assume that LoRA weights are initialized with \init{1} or \init{2} and trained with gradient descent with learning rate $\eta = \Theta(n^c)$ for some $c \in \reals$. Then, it is impossible to have $\delta_t^i = \Theta(1)$ for $i \in \{1,2\}$ for any $t>0$, and therefore, fine-tuning with LoRA in this setup is inefficient.
\end{prop}
In conclusion, efficiency cannot be achieved with this parametrization of the learning rate. This suggests that standard LoRA finetuning as currently used by practitioners is suboptimal, especially when model width is large, which is a property that is largely satsified in practice ($n\approx 700$ for GPT2 and $n\approx 4000$ for LLama). This analysis suggests that \emph{we are missing crucial hyperparameters} in the standard LoRA setup. Indeed, we show that by decoupling the learning rate for $a$ and $b$, we can have $\delta_t^i = \Theta(1)$ for $i \in \{1,2, 3\}$. We write $\eta_a, \eta_b$ to denote the learning rates. The analysis conducted above remains morally the same with the only difference being in the learning rates. Let $\eta_a = \Theta(n^{c_a})$ and $\eta_b = \Theta(n^{c_b})$, and assume that weights are initialized with \init{1}. A similar analysis to the one conducted above show that having $f_t(x) = \Theta(1)$ and $\delta_t^i = \Theta(1)$ for $i \in \{1,2\}$ and $t>0$ implies that for all $t>1$
\begin{equation*}
    \begin{cases}
c_a+2\gamma[b_{t-1}] + 1 = 0 \quad (\delta^1_t = \Theta(1))\\
        c_b + 2 \gamma[a_{t-1}^\top x] = 0\quad (\delta^2_t = \Theta(1)) \\
        \gamma[b_{t-1}] + \gamma[a_{t-1}^\top x] = 0 \quad (f_{t-1}(x) = \Theta(1))
    \end{cases}
\end{equation*}

which, after simple calculations, implies that $c_a + c_b = -1$. This is only a necessary condition. In the next result, taking also some elements of stability into consideration, we fully characterize the choice of $\eta_a$ and $\eta_b$ to ensure efficient LoRA fine-tuning. 

\begin{prop}[Efficient Fine-Tuning with LoRA]\label{prop:efficient_toy}
In the case of model \eqref{eq:toy_model}, with $\eta_a = \Theta(n^{-1})$ and $\eta_b = \Theta(1)$, we have for all $t>1$, $i \in \{1,2,3\}$, $\delta_t^i = \Theta(1)$.
\end{prop}

We refer the reader to \cref{app:proofs} for more details on the proof of \cref{prop:efficient_toy}.  In conclusion, scaling the learning rates as $\eta_a = \Theta(n^{-1})$ and $\eta_b = \Theta(1)$ ensures stability ($\Delta f_t = \Theta(1)$) and efficiency of LoRA finetuning ($\delta^i_t = \Theta(1)$ for $i\in\{1,2\}$ and $t>1$) in the infinite-width limit. In practice, this means that the learning rate for $b$ should be generally much larger than that of $a$. This remains true even if $b \in \reals^{r}$ for general $r$. We will later see that this scaling is valid for general neural network models. 
\begin{figure}[h]
    \centering
     \includegraphics[width=.95\linewidth]{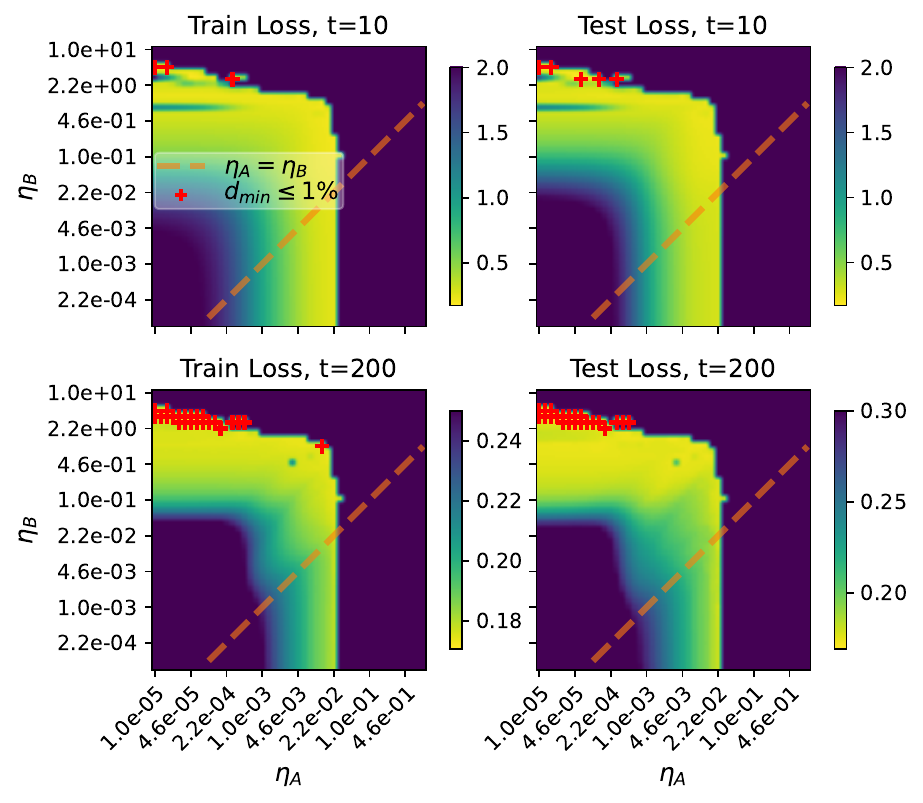}
    \includegraphics[width=0.91\linewidth]{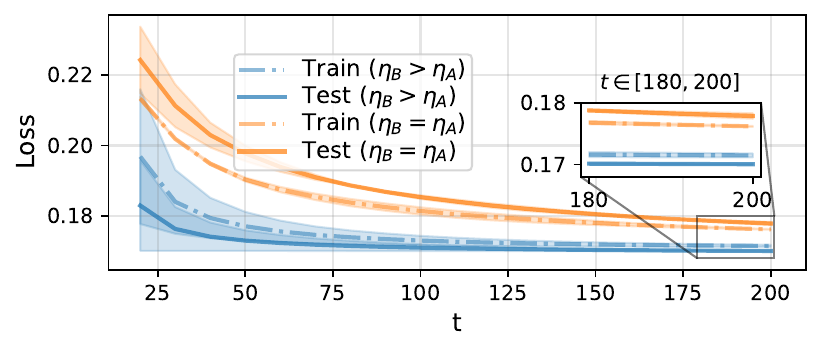}
    \caption{\small{(\textbf{Top}) Train/Test accuracy of toy model \cref{eq:toy_model_exps} averaged over 3 random seeds. Orange dashed line represents the line $\eta_A = \eta_B$, and red dots represents all values of $(\eta_A, \eta_B)$ for which $d_{\min}(\eta_A, \eta_B):= \loss_{(\eta_A,\eta_B)}/\loss^* - 1 \leq 1\%$, where $\loss^*$ is the best loss. (\textbf{Bottom}) Train/Test curves for two sets of learning rates: the optimal choice $(\eta_A^*, \eta_B^*) = (2.78, 1.29\mathrm{e}{-4})$ overall at $t=200$ in terms of test loss (Blue) and the optimal choice when $\eta_A =  \eta_B$ which is given by $(\eta_A, \eta_B) = (2.15\mathrm{e}{-2}, 2.15\mathrm{e}{-2})$ (Orange). All values are averaged oevr three runs and confidence interval are shown (shaded).}}
    \vspace{-0.4cm}
    \label{fig:toy_heatmap}
\end{figure}

\subsection{Verifying the Results on a Toy Model}
The previous analysis considers a simple linear model. To assess the validity of the scaling rules in a non-linear setting, we consider a neural network model given by 
\begin{equation}\label{eq:toy_model_exps}
f(x) = W_{out} \phi(BA \phi(W_{in} x)),
\end{equation}
where $W_{in} \in \reals^{n\times d}, W_{out} \in \reals^{1\times n}, A \in \reals^{r\times n}, B \in \reals^{n \times r}$ are the weights, and $\phi$ is the ReLU function.  
The model is trained on a synthetic dataset generated with $X\sim \normal(0, I_d), \,\, Y = \sin(d^{-1}\sum_{i=1}^d X_i)$. 
See \cref{app:add_exps} for more details.

Only the weight matrices $A, B$ are trained ($W_{in}, W_{out}$ are fixed). We use $d=5, n=100, r=4$, train data size $1000$ and a test data size $100$.\footnote{See \cref{app:add_exps} for more details about the experimental setup.} The train/test loss for varying $\eta_A$ and $\eta_B$ is reported in \cref{fig:toy_heatmap} at the early stages of the training ($t=10$) and after convergence (we observed convergence around $t \approx 200$ for reasonable choices of learning rates). The red '$+$' signs represents learning rates $(\eta_A, \eta_B)$ for which the loss is within $1\%$ range from the best loss and dashed line represents the case where the learning rates are set equal. We observe that both the best train and test losses are consistently achieved by a combination of learning rates where $\eta_b \gg \eta_a$, which validates our analysis in the previous section. Notice also that optimal learning rates $(\eta_A, \eta_B)$ are generally close to the edge of stability, a well-known behaviour in training dynamics of deep networks \citep{cohen2021gradient}.

\section{Stability and Feature Learning with LoRA in the Infinite Width Limit}\label{sec:main_theory}

In this section, we extend the analysis above to general neural architectures with LoRA layers. We show that the conclusions from the analysis on the linear model hold for general neural architectures: 1) using the same learning rate for both $A$ and $B$ leads to suboptimal feature learning when model width is large, and  2) this problem can be fixed by setting different learning rates for $A$ and $B$. 

Since our aim in this paper is primarily methodological, the theoretical results in this section are of a physics level of rigor, omitting technical assumptions that would otherwise make the analysis rigorous but unnecessarily complicated. In all the results, LoRA rank $r$ is considered fixed and finetuning dynamics are analyzed in the limit of infinite-width. This setup fairly represents practical scenarios where $r \ll n$ and $r$ is generally small.
\vspace{-0.3cm}
\paragraph{Notation.} The LoRA weights are initialized with $A_{ij} \sim \normal(0,\sigma_A^2), B_{ij} \sim \normal(0,\sigma_B^2)$ for some $\sigma_A, \sigma_B \geq 0$.\footnote{In \cite{hu2021lora}, $B$ is initialized to $0$, which corresponds to setting $\sigma_B = 0$.} Here also, we assume that either $\sigma_B^2 = 0$ and $\sigma_A^2 = \Theta(n^{-1})$ (\init{1}), or $\sigma_B^2 = \Theta(1)$ and $\sigma_A^2 = 0$ (\init{2}). Given a LoRA layer in the model, $\underbar{Z}$ denotes the input to that layer and $\Bar{Z}$ the output after adding the pretrained weights. More precisely, we write $\Bar{Z} = W^*\underbar{Z} + \frac{\alpha}{r}BA\,\underbar{Z}$.

Our main analysis relies on a careful estimation of the magnitude of several quantities including \emph{LoRA features}. Let us first give a formal definition.
\begin{defn}[LoRA Features]\label{def:lora_features}
Given a general neural architecture and a LoRA layer (\cref{def:lora}), we define LoRA features $(Z_A, Z_B)$ as $Z_A = A \underbar{Z}$ and $Z_B = B Z_A = BA \underbar{Z}$\,.
At fine-tuning step $t$, we use the superscript $t$ to denote the value of LoRA features $Z_A^t, Z_B^t$, and the subscript $t$ to denote the weights $A_t, B_t$. 
\end{defn}

LoRA layers are 2-layers linear networks with a ``bottleneck'' in the middle (since generally $r\ll n$). This bottleneck shape might induce some numerical challenges in training stability and efficiency (\cref{def:stability} and \cref{def:efficient_learning}). 
\vspace{-0.3cm}
\paragraph{Finetuning Dataset.} To simplify the analysis, we assume that the finetuning dataset comprises a single sample $(x,y)$,\footnote{This assumption on the finetuning dataset is for simplification purposes only. All our analysis can be re-written with `batched' gradients and the conclusions remain the same. However, some additonal assumptions are required to make the analysis rigorous.} and the goal is to minimize the loss $\loss(\bm{\theta},(x,y))$ computed with the underlying model where the adjusted weights are given by $W^{*} + BA$ for all LoRA layers (here $\bm{\theta} = \{A, B, \textrm{ for all LoRA layers in the model}\}$). At training step $t$, and for any LoRA layer in the model, $\underbar{Z}^t$ is the input to the LoRA layer, computed with data input $x$. Similarly, we write $d \Bar{Z}^t$ to denote the gradient of the loss function with respect to the layer output features $\Bar{Z}$ evaluated at data point $(x,y)$. 

The notion of stability of LoRA as discussed in \cref{sec:toy} can be generalized to any neural network model as follows.
\begin{defn}[Stability]\label{def:stability} 
We say that LoRA finetuning is stable if for all LoRA layers in the model, and all training steps $t$, we have $\underbar{Z}, Z_A, Z_B = \bigO(1)$ as $n$ goes to infinity.
\end{defn}
Stability implies that no quantity in the network explodes as width grows, a desirable property as we scale the model.\footnote{It is possible to define stability as $\underbar{Z}, Z_B=\bigO(1)$ and exclude $Z_A$ from the condition. This would allow scenarios where for instance the entries of $A$ explode with width but their magnitude is compensated with a smaller magnitude of $B$. This system has one degree of freedom because of the homogeneity of the product $BA$, and by imposing that $Z_A=\bigO(1)$, we avoid having such scenarios.} Naturally, in order to ensure stability, one has to scale hyperparameters (initialization, learning rate) as $n$ grows. Scaling rules for initialization are fairly easy to infer and were already discussed in \cref{sec:toy} where we obtained two plausible initialization schemes (\init{1} and \init{2}). More importantly, if we arbitrarily scale the learning rate with width, we might end up with suboptimal learning as width grows even if the finetuning is stable. This is the case for instance when we aggressively downscale the learning rate with width, or inadequately parameterize the network (e.g. Neural Tangent Kernel parametrization which leads to the kernel regime in the infinite width limit, \cite{jacot2018neural}).
To take this into account, we define a notion of feature learning with LoRA.
\begin{defn}[Stable Feature Learning with LoRA]\label{def:stable_feature_learning_lora} 
We say that LoRA finetuning induces stable feature
learning if it is stable (\cref{def:stability}), and for all LoRA layers and finetuning step $t$, we have $\Delta Z_B^t \overset{def}{=} Z_B^{t+1} - Z_B^t = \Theta(1)$.
\end{defn}
A similar definition of feature learning was introduced in \cite{yang2023tensor} for pretraining. This definition ensures that the network is not `stuck' in a kernel regime where feature updates are of order $\bigO(n^{-\epsilon})$  in the infinite-width limit for some $\epsilon > 0$, which implies that no feature learning occurs in the limit. The authors introduced the $\mu$-parameterization (or maximal update parametrization), a specific network parameterization (initialization + learning rate scaling), that ensures that feature updates are $\Theta(1)$.
Note that here we added stability in the definition, but in principle, one could define feature learning with $\Omega$ instead of $\Theta$. The latter covers unstable scenarios (e.g. when $\Delta Z_B^t = \Theta(n)$ due to improper scaling of initialization and learning rate), so we omit it here and focus on stable feature learning. Also, notice that we only consider finetuning dynamics and not the pretraining dynamics. However, since our analysis depends on weights $W^*$ from pretraining, we assume that pretraining parameterization ensures stability and feature learning as width grows (see \cref{app:proofs} for more details).\footnote{When taking the infinite width limit, we assume that pretraining parameterization is $\mu$P. This is just a technicality for the infinite-width limit and does not have any implications on practical scenarios where the width is finite. The most important implications of this assumption is that in the pretrained network (before introducing LoRA layers), we have $\underbar{Z} = \Theta(1), \Bar{Z} = \Theta(1)$, which holds for a general input-output pair $(x,y)$.}

At finetuning step $t$, the gradients are given by
\begin{align*}
\frac{\partial \loss_t}{\partial B} &= \frac{\alpha}{r} d\Bar{Z}^{t-1} \otimes A_{t-1} \underbar{Z}^{t-1}\\
\frac{\partial \loss_t}{\partial A} &= dZ_A^{t-1} \otimes \underbar{Z}^{t-1} = \frac{\alpha}{r} B^\top_{t-1} d\Bar{Z}^{t-1} \otimes \underbar{Z}^{t-1},
\end{align*}
where $u \otimes v$ denotes the outer product $uv^\top$ of vectors $u$, $v$,
and the weights are updated as follows
$$
A_{t} = A_{t-1} - \eta_A g_A^{t-1}, \quad B_{t} = B_{t-1} - \eta_B g_B^{t-1},
$$
where $g_A, g_B$ are processed gradients (e.g. normalized gradients with momentum as in AdamW etc). Hereafter, we assume that the gradients are processed in a way that makes their entries $\Theta(1)$. This is generally satisfied in practice (with Adam for instance) and has been considered in \cite{yang2023tensor} to derive the $\mu$-parametrization for general gradient processing functions. 

Unlike the linear model in \cref{sec:toy}, LoRA feature updates are not only driven by the change in the $A, B$ weights, but also $\underbar{Z}, d\bar{Z}$ which are updated as we finetune the model (assuming there are multiple LoRA layers). To isolate the contribution of individual LoRA layers to feature learning, we assume that only a \emph{single LoRA layer is trainable} and all other LoRA layers are frozen.\footnote{This is equivalent to having only a single LoRA layer in the model since LoRA layers are initialized to zero. In this way, we can quantify feature learning induced by the LoRA layer as we finetune the model.}. In this setting, considering the only trainable LoRA layer in the model, the layer input $\underbar{Z}$ is fixed and does not change with $t$, while $d\bar{Z}$ changes with step $t$ (because $\bar{Z}^t = (W^* + \frac{\alpha}{r}B_tA_t)\underbar{Z}$). After step $t$, $Z_B$ is updated as follows 
$$
\Delta Z_B^t = \underset{\delta_t^1}{\underbrace{B_{t-1} \Delta Z_A^t }} + \underset{\delta_t^2}{\underbrace{\Delta B_t Z_A^{t-1}}} + \underset{\delta^3_t}{\underbrace{\Delta B_t \Delta Z_A^t}}
$$
As discussed in \cref{sec:toy}, the terms $\delta^1_t, \delta^2_t$ represent the `linear' feature updates that we obtain if we fix one weight matrix and only train the other, while $\delta^3_t$ represents the `multiplicative' feature update which captures the compounded update due to updating both $A$ and $B$. 

\paragraph{Analysis of the Role of $A$ and $B$.} As discussed above,  we want to ensure that $\delta_t^1 = \Theta(1)$ and $\delta_t^2 = \Theta(1)$ which means that both weight matrices contribute to the update in $Z_B$. To further explain why this is a desirable property, let us analyze how changes in matrices $A$ and $B$ affect LoRA feature $Z_B = BA \,\underbar{Z}$. 

Let $(B_{:,i})_{1\leq i \leq r}$ denote the columns of $B$. We can express $Z_B$ as
$
Z_B = \sum_{i=1}^r (A \, \underbar{Z})_i B_{:,i}
$, where $(A\underbar{Z})_i$ is the $i^{th}$ coordinate of $A \underbar{Z}$. This decomposition suggests that the \emph{direction} of $Z_B$ is a weighted sum of the columns of $B$, and $A$ modulates the \emph{weights}. With this, we can also write  
\begin{equation*}
    \begin{cases}
        \delta^1_t = \sum_{i=1}^r (\Delta A_t \underbar{Z})_i (B_{:,i})_{t-1}\\
        \delta^2_t  = \sum_{i=1}^r ( A_{t-1} \underbar{Z})_i (\Delta B_{:,i})_{t-1},
    \end{cases}
\end{equation*}
where $(B_{:,i})_{t}$ refers to the columns of $B$ at time step $t$. Having both $\delta_t^1$ and $\delta_t^2$ of order $\Theta(1)$ means that both $A$ and $B$ are `sufficiently' updated to induce a change in weights $(A \underbar{Z})_i$ and directions $B_{:,i}$.
If one of the matrices $A, B$ is not efficiently updated, we might end up with suboptimal finetuning, leading to either non updated directions $B$ or direction weights $(A_{t-1}Z)$. For instance, assuming that the model is initialized with \init{2}, and that $B$ is not efficiently updated, the direction of $Z_B$ will be mostly determined by the vector (sub)space of dimension $r$ generated by the columns of $B$ at initialization. This analysis leads to the following definition of efficient learning with LoRA.
\begin{figure*}
    \centering
    \includegraphics[width=0.90\linewidth]{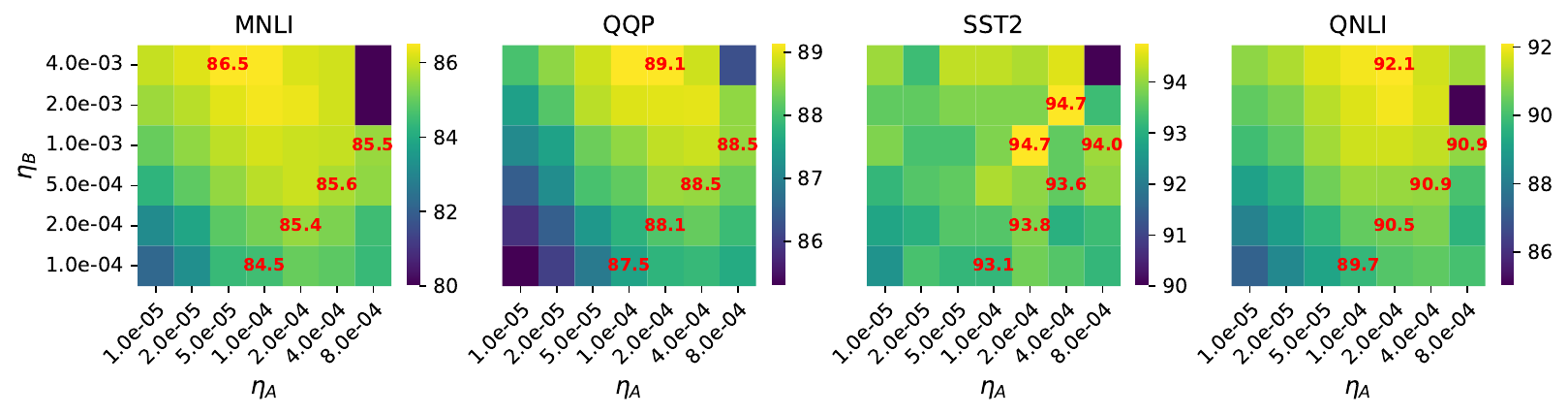}
    \caption{\small{Test accuracy of Roberta-base finetuning for $3$ epochs on MNLI, QQP, QNLI, and $10$ epochs on SST2, with sequence length $T=128$ and half precision (FP16). LoRA hyperparameters are set to $\alpha=r=8$. All values are averaged over 3 random seeds (we do not show confidence intervals for better visualizations, but fluctuations are of order $0.1\%$, see \cref{fig:comparison_standard_vs_ours} for instance). For better visualization, when accuracy is lower than a fixed threshold, we set it to threshold. Values shown in red are:  1) the best accuracy (overall) and 2) the accuracy for a set of learning rates where $\eta_B$ and $\eta_A$ are close in order of magnitude ($\eta_B/\eta_A \in [1,1.25]$)}.}
    \label{fig:roberta_main}
\end{figure*}
\begin{defn}[Efficient Learning]\label{def:efficient_learning}
    We say that LoRA fine-tuning is efficient if it is stable (\cref{def:stability}), and for all LoRA layers in the model, all steps $t>1$, and $i \{1,2\}$, we have 
    $
    \delta_t^i = \Theta(1)
    $.
\end{defn}
Note that it is possible to achieve stable feature learning (\cref{def:stable_feature_learning_lora}) without necessarily having efficient learning. This is the case when for instance $B$ is not updated (fixed to a non-zero init with \init{2}) and only $A$ is updated, which corresponds to simply setting $\eta_B=0$. This is a trivial case, but other non-trivial cases of inefficiency are common in practice, such as the use of the same learning rate for $A$ and $B$ which is a standard practice. In the next theorem, we characterize the optimal scaling of learning rates $\eta_A$ and $\eta_B$, a conclusion similar to that of \cref{sec:toy}.

\begin{thm}[Efficient LoRA (Informal)]\label{thm:efficiency}
Assume that weight matrices $A$ and $B$ are trained with Adam with respective learning rates $\eta_A$ and $\eta_B$. Then, it is impossible to achieve efficiency with $\eta_A = \eta_B$. However, LoRA Finetuning is efficient with $\eta_A = \Theta(n^{-1})$ and $\eta_B = \Theta(1)$.
\end{thm}

The result of \cref{thm:efficiency} suggests that efficiency can only be achieved with $\eta_B / \eta_A = \Theta(n)$. In practice, this translates to setting $\eta_B \gg \eta_A$, but does not provide a precise ratio $\eta_B/\eta_A$ to be fixed while tuning the learning rate (the constant in `$\Theta$' is generally intractable), unless we tune both $\eta_B$ and $\eta_A$ which is not efficient from a computational perspective as it becomes a 2D tuning problem. It is therefore natural to set a fixed ratio $\eta_B/\eta_A$ and tune only $\eta_A$ (or $\eta_B$), which would effectively reduce the tuning process to a 1D grid search, achieving the same computational cost of standard LoRA where the learning rate is the same for $A$ and $B$. We call this method LoRA$+$.

\newtcolorbox{mybox}[2][]{colbacktitle=red!10!white,
colback=green!15!white,coltitle=red!70!black,
title={#2},fonttitle=\bfseries,#1}
\begin{mybox}[attach title to upper={\  :\ }]{LoRA$+$}
set the learning rates for $A, B$ such that $\eta_B = \lambda \eta_A$ with $\lambda>1$ fixed and tune $\eta_A$.
\end{mybox}
In the next section, through extensive empirical evaluations, we first validate our theoretical result and show that optimal pairs $(\eta_A,\eta_B)$ (in terms of test accuracy) generally satisfy $\eta_B \gg \eta_A$. We then investigate the optimal ratio $\lambda$ for LoRA$+$ and suggest a default ratio that was empirically found to generally improve performance compared to standard LoRA. Although the conclusions of \cref{thm:efficiency} and \cref{prop:efficient_toy} are similar, the proof techniques are different. In \cref{prop:efficient_toy}, the linear model is trained with gradient descent, while in \cref{thm:efficiency}, the training algorithm is Adam-type in the sense that it normalizes the gradients before updating the weights. The formal statement of \cref{thm:efficiency} requires an additional assumption on the alignment of the processed gradients $g_A$ with LoRA input $\underbar{Z}$. This technical detail is introduced and discussed in \cref{app:proofs}.

\section{Experiments with Language Models}
We report our empirical results using LoRA to finetune a set of language models on different benchmarks. Details about the experimental setup and more empirical results are provided in \cref{app:add_exps}. We also identify a default value for the ratio $\lambda = \eta_B / \eta_A$ that generally improves performance as compared to standard LoRA. The code for our experiments is available at \url{https://github.com/nikhil-ghosh-berkeley/loraplus}.
\subsection{GLUE tasks with GPT-2 and RoBERTa}
\begin{figure}[h]
    \centering
    \includegraphics[width=0.94\linewidth]{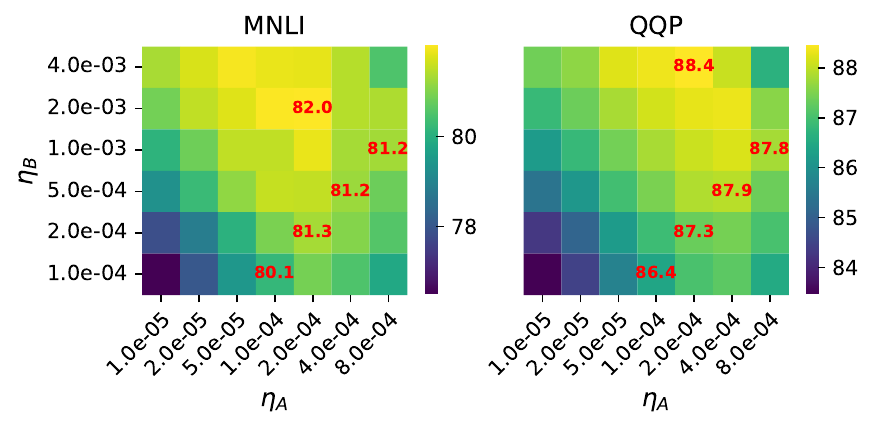}
    \vspace{-0.2cm}
    \caption{\small{Test accuracy of GPT-2 after finetuning for $3$ epochs on MNLI, QQP, with FP16 precision. LoRA hyperparameters are set to $\alpha=r=8$. Both train/test accuracy are consistently maximal for some choice of learning rates where $\eta_B \gg \eta_A$. See \cref{app:add_exps} for more numerical results with GPT2.}}
    \label{fig:gpt2_main}
    \vspace{-0.6cm}
\end{figure}

The GLUE benchmark (General Language Understanding Evaluation) consists of several language tasks that evaluate the understanding capabilities of langugage models \citep{wang2018glue}. Using LoRA, we finetune Roberta-base from the RoBERTa family \citep{liu2019roberta} and GPT-2 \citep{radford2019language} on MNLI, QQP, SST2, and QNLI tasks (Other tasks are smaller and generally require an already finetuned model e.g. on MNLI as starting checkpoint) with varying learning rates $(\eta_A, \eta_B)$ to identify the optimal combination. Empirical details are provided in \cref{app:add_exps}. 
\vspace{-0.32cm}
\paragraph{Roberta-base.} \cref{fig:roberta_main} shows the results of Roberta-base finetuning with $\alpha = r=8$, trained with half precision (FP16). We observe that test accuracy is consistently maximal for some set of learning rates satisfying $\eta_B \gg \eta_A$, outperforming the standard practice where $\eta_A$ and $\eta_B$ are usually set equal. Interestingly, the gap between the optimal choice of learning rates overall and the optimal choice when $\eta_A \approx \eta_B$ is more pronounced for `harder' tasks like MNLI and QQP, as compared to SST2 and QNLI. This is probably due to the fact that harder tasks require more efficient feature learning. It is also worth mentioning that in our experiments, given limited computational resources, we use sequence length $T=128$ and finetune for only $3$ epochs for MNLI and QQP, so it is expected that we obtain test accuracies lower that those reported in \cite{hu2021lora} where the authores finetune Roberta-base with $T=512$ sequence length (for MNLI) and more epochs ($30$ for MNLI). In \cref{app:add_exps}, we provide additional results with Test/Train accuracy/loss.
\paragraph{GPT-2.} \cref{fig:gpt2_main} shows the results of finetuning GPT-2 with LoRA on MNLI and QQP (other tasks and full precision training are provided in \cref{app:add_exps}). Similar to the conclusions from Roberta-base, we observe that maximal test accuracies are achieved with some $(\eta_A, \eta_B)$ satisfying $\eta_B \gg \eta_A$. Further GPT-2 results with different tasks are provided in \cref{app:add_exps}. Here also, we observed that the harder the task, the larger the gap between model performance when $\eta_B \gg \eta_A$ and when $\eta_A \approx \eta_B$.
\subsection{Llama}
To further validate our theoretical findings, we finetune the Llama-7b model \citep{touvron2023llama} on the MNLI dataset and flan-v2 dataset \citep{longpre2023flan} using LoRA. Each trial is averaged over two seeds.
\paragraph{Flan-v2.}
We examine LoRA training of Llama on the instruction finetuning dataset flan-v2 \citep{longpre2023flan}. To make the experiments computationally feasible, we train for one epoch on a size $100,000$ subset of the flan-v2 dataset. We record the test accuracy of the best checkpoint every 500 steps. The LoRA hyperparameters are set to $\alpha = 16$ and $r = 64$. The adapters are added to every linear layer (excluding embedding layers) and we use a constant learning rate schedule. The full training details are in \cref{app:add_exps}. 
\begin{figure}[h]
    \centering
    \includegraphics[width=0.94\linewidth]{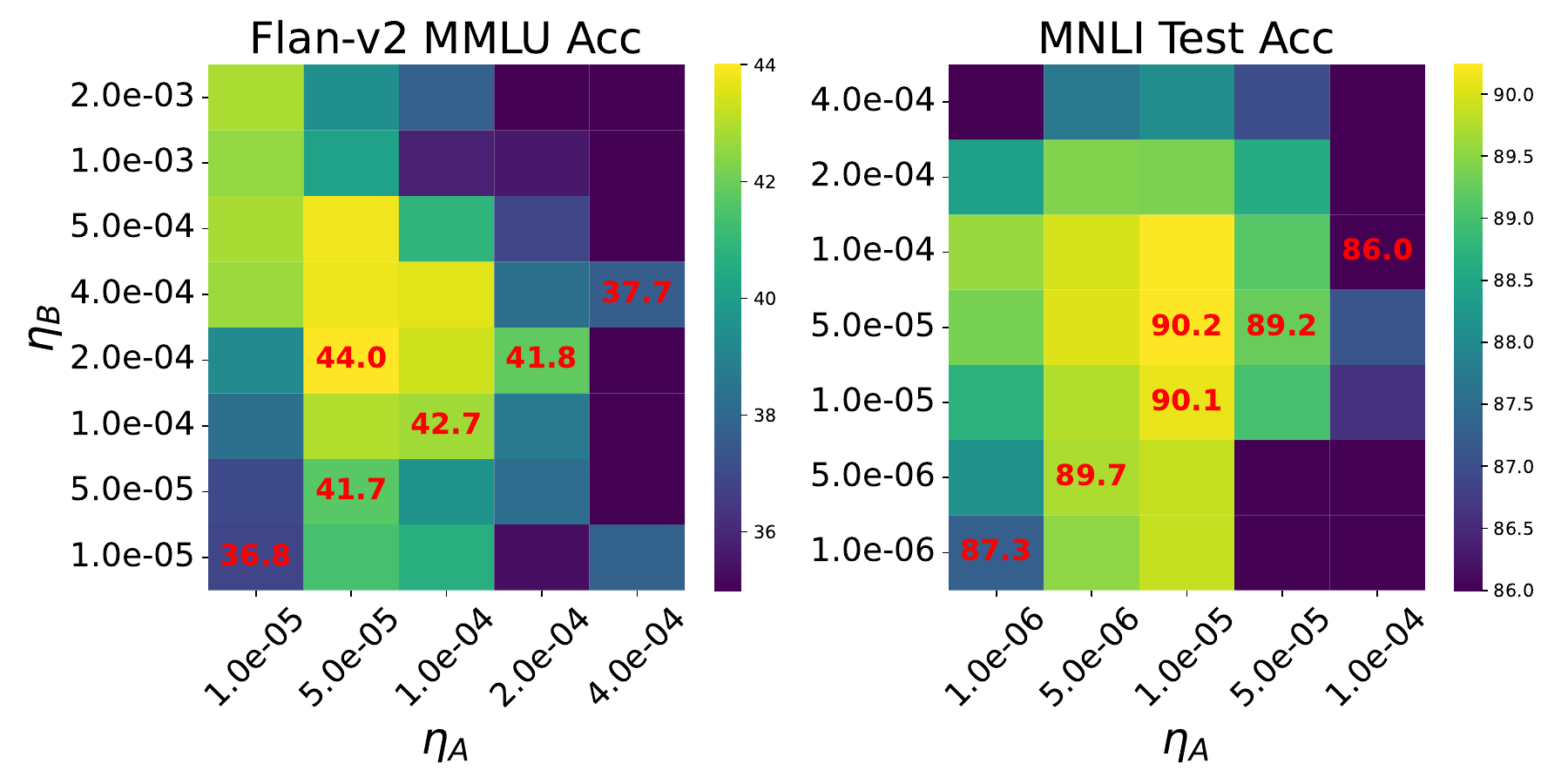}
    \vspace{-0.2cm}
    \caption{\small{Left: MMLU accuracy of Llama-7b trained for one epoch on a 100k subset of flan-v2. Right: Test accuracy of the best checkpoint of Llama-7b trained on MNLI for one epoch. Values are averaged over two seeds.}}
    \vspace{-0.2cm}
    \label{fig:llama_test}
\end{figure}
  
We evaluate the final model on the MMLU benchmark \citep{hendrycks2020measuring}. The results in Figure \ref{fig:llama_test} show that for this benchmark taking $\eta_B \gg \eta_A$ is advantageous and results in a roughly 1.3\% gain compared with the optimal $\eta_B = \eta_A$. In \cref{app:add_exps} we show that the same effect holds also when using \init{1}. 

\paragraph{MNLI.}  The right panel of Fig \ref{fig:llama_test} shows the results of finetuning Llama-7b with LoRA on MNLI, with $\alpha = 16$, $r = 8$. We train using half precision and constant learning rate schedule, with a sequence length $T=128$. Since MNLI is relatively easy for Llama, we finetune for only one epoch, which is sufficient for the model to reach its peak test accuracy. In Figure \ref{fig:llama_test}, $\eta_B = \eta_A$ is nearly optimal for all $\eta_B \geq \eta_A$. This is consistent with the intuition that efficient feature learning is not required for easy tasks and that having $\eta_B / \eta_A \gg 1$ does not significantly enhance performance. Additionally, the magnitude of stable learning rates for Llama is much smaller than for GPT-2 and RoBERTa on MNLI further supporting that Llama requires less adaptation. Analogous plots for the train and test loss are shown in Fig  \ref{fig:llama_mnli_loss} in \cref{app:add_exps}.

\subsection{How to set LoRA+ Ratio?}

\begin{figure}[h]
    \centering
    \includegraphics[width=0.9\linewidth]{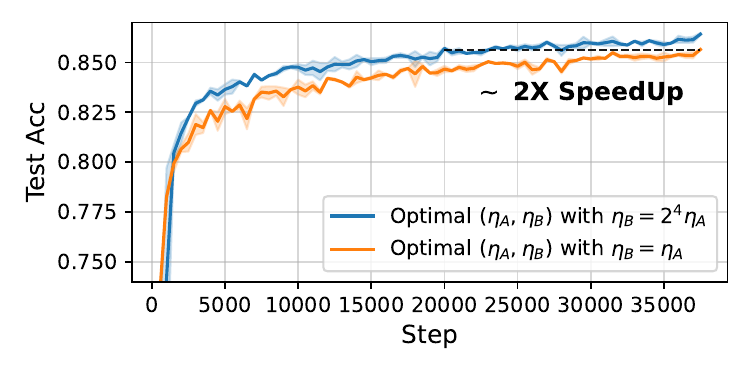}
    \vspace{-0.2cm}
    \caption{\small{Test accuracy of Roberta-base finetuned on the MNLI task in two setups: (\textbf{LoRA+}) $\eta_B = 2^4 \eta_A$ and (\textbf{Standard}) $\eta_B = \eta_A$. $\eta_A$ is tuned using a grid search.}}
    \vspace{-0.2cm}
    \label{fig:comparison_standard_vs_ours}
\end{figure}

Naturally, the optimal ratio $\lambda$ depends on the architecture and the finetuning task via the constants in `$\Theta$' (\cref{thm:efficiency}). This is a limitation of these asymptotic results since they do not offer any insights on how the constants are affected by the task and the neural architecture. 
\begin{figure}[h]
    \centering
    \includegraphics[width=0.9\linewidth]{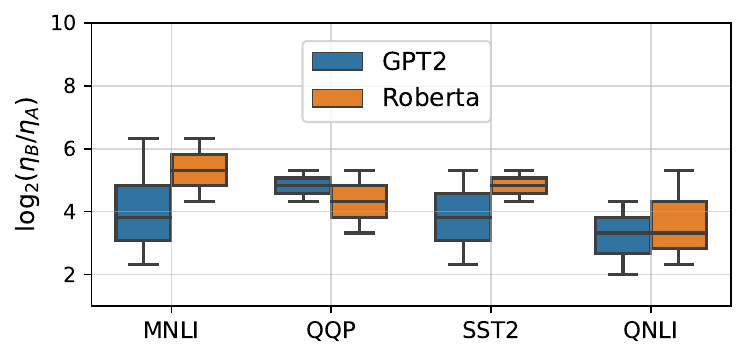}
    \vspace{-0.2cm}
    \caption{\small{Distribution of the ratio $\eta_B/\eta_A$ for the top 4 learning rate for each pair (model, task). The 4 learning rates are selected using the test loss at the end of finetuning (i.e. top 4 learning rates $(\eta_B, \eta_A)$ in terms of test loss). The distribution shows the interquartile range (
    $25\% - 75\%$ quantiles) and the median. }}
    \vspace{-0.2cm}
    \label{fig:optimal_ratio}
\end{figure}
\cref{fig:optimal_ratio} show the distribution of the ratio $\eta_B/\eta_A$ for the top $4$ runs in terms of test accuracy for different pairs of (model, task). This is the same experimental setup of \cref{fig:roberta_main} and \cref{fig:gpt2_main}. The optimal ratio is model and task sensitive and shows significant variance. Our additional experiments in \cref{app:add_exps} show that it is also sensitive to initialization (\init{1} vs \init{2}). With \init{2}, we found that generally setting a ratio of $\lambda = \eta_B / \eta_A \approx 2^4$ improves performance for Roberta (\cref{fig:comparison_standard_vs_ours}). However, with \init{1}, we found that the optimal ratio is smaller and is of order $2^2$-$2^3$ (see \cref{app:add_exps}). For LLama experiments, it seems that a ratio of order $2^1$-$2^2$ is optimal..

\section{Conclusion and Limitations}

Employing a scaling argument, we showed that LoRA finetuning as it is currently used in practice is not efficient. We proposed a method, LoRA+, that resolves this issue by setting different learning rates for LoRA adapter matrices. Our analysis is supported by extensive empirical results confirming the benefits of LoRA+ for both training speed and performance. These benefits are more significant for `hard' tasks such as MNLI for Roberta/GPT2 (compared to SST2 for instance) and MMLU for LLama-7b (compared to MNLI for instance).
However, as we depicted in \cref{fig:comparison_standard_vs_ours}, a more refined estimation of the optimal ratio $\eta_B/\eta_A$ should take into account task and model dependent, and our analysis in this paper lacks this dimension. We leave this for future work.

\newpage

\section*{Acknowledgement}
We thank Amazon Web Services (AWS) for cloud credits under an Amazon Research Award. 
We also gratefully acknowledge partial support from NSF grants DMS-2209975, 2015341, NSF grant 2023505 on Collaborative Research: Foundations of Data Science Institute (FODSI), the NSF and the Simons Foundation for the Collaboration on the Theoretical Foundations of Deep Learning through awards DMS-2031883 and 814639, and NSF grant MC2378 to the Institute for Artificial CyberThreat Intelligence and OperatioN (ACTION).

\section*{Impact Statement}
This paper presents work whose goal is to advance the field of Machine Learning, specifically, to speed up the leading algorithm LoRA for fine-tuning pre-trained large language models while improving performance of the fine-tuned models.
The speed-up saves computation resources when pre-trained large language models are customized for particular down-stream tasks.
There are many potential societal consequences of our work, none which we feel must be specifically highlighted here.

\bibliography{ref}

\newpage
\onecolumn
\appendix

\section{Proofs}\label{app:proofs}
In this section, we provide proofs for \cref{prop:inefficiency_toy}, \cref{prop:efficient_toy}, \cref{thm:efficiency}, and some technical details used in the proofs.

\subsection{Scaling of Neural Networks}\label{sec:scaling}
Scaling refers to the process of increasing the size of one of the ingredients in the model to improve performance (see e.g. \cite{hoffmann2022training}). This includes model capacity which can be increased via width (embedding dimension) or depth (number of layers) or both, compute (training data), number of training steps etc. In this paper, we are interested in scaling model capacity via the width $n$. This is motivated by the fact that most state-of-the-art language and vision models have large width.

It is well known that as the width $n$ grows, the network initialization scheme and the learning should be adapted to avoid numerical instabilities and ensure efficient learning. For instance, the initialization variance should scale $1/n$ to prevent arbitrarily large pre-activations as we increase model width $n$ (e.g. He init \cite{he2016deep}). To derive such scaling rules, a principled approach consist of analyzing statistical properties of key quantities in the model (e.g. pre-activations) as $n$ grows and then adjust the initialization, the learning rate, and the architecture itself to achieve desirable properties in the limit $n \to \infty$ \cite{hayou19activation, deepinfoprop2017, yang2019scaling}.

In this context, \cite{yang2022tensor} introduces the Maximal Update Parameterization (or $\mu$P), a set of scaling rules for the initialization scheme, the learning rate, and the network architecture that ensure stability and maximal feature learning in the infinite width limit. Stability is defined by $Y_l^i = \Theta(1)$ for all $l$ and $i$ where the asymptotic notation `$\Theta(.)$' is with respect to width $n$ (see next paragraph for a formal definition), and feature learning is defined by $\Delta Y_l = \Theta(1)$, where $\Delta$ refers to the feature update after taking a gradient step. $\mu$P guarantees that these two conditions are satisfied at any training step $t$. Roughly speaking, $\mu$P specifies that hidden weights should be initialized with $\Theta(n^{-1/2})$ random weights, and weight updates should be of order $\Theta(n^{-1})$. Input weights should be initialized $\Theta(1)$ and the weights update should be $\Theta(1)$ as well. While the output weights should be initialized $\Theta(n^{-1})$ and updated with $\Theta(n^{-1})$. These rules ensure both stability and feature learning in the infinite-width limit, in contrast to standard parameterization (exploding features if the learning rate is well tuned), and kernel parameterizations (e.g. Neural Tangent Kernel parameterization where $\Delta Y_l = \Theta(n^{-1/2})$, i.e. no feature learning in the limit). 

\subsection{The Gamma Function \texorpdfstring{($\gamma[.]$)}{Gamma Function}}\label{sec:gamma_fct}

In the theory of scaling of neural networks, one usually tracks the asymptotic behaviour of key quantities as we scale some model ingredient. For instance, if we scale the width, we are interested in quantifying how certain quantities in the network behave as width $n$ grows large and the asymptotic notation becomes natural in this case. This is a standard approach for (principled) model scaling and it has so far been used to derive scaling rules for initialization \citep{deepinfoprop2017}, activation function \citep{hayou19activation}, network parametrization \citep{yang2023depth}, amongst other things.

With \init{1} and \init{2}, the weights are initialized with $\Theta(n^{-\beta})$ for some $\beta \geq 0$. Assuming that the learning rates also scale polynomially with $n$, it is straightforward that preactivations, gradients, and weight updates are all asymptotically polynomial in $n$. It is therefore natural to introduce the Gamma function, and we write $v = \Theta(\gamma[v])$ to capture this polynomial behaviour. Now, let us introduce some elementary operations with the Gamma function.

\paragraph{Multiplication.} Given two real-valued variables $v,v'$, we  have $\gamma[v \times v'] = \gamma[v] + \gamma[v']$. 
\paragraph{Addition.} Given two real-valued variables $v,v'$, we generally have $\gamma[v + v'] = \max(\gamma[v], \gamma[v'])$. The only case where this is violated is when $v' = -v$. This is generally a zero probability event if $v$ and $v'$ are random variables that are not perfectly correlated, which is the case in most situations where we make use of this formula (see the proofs below). 

\subsection{Proof of \texorpdfstring{\cref{prop:inefficiency_toy}}{Proposition Inefficiency Toy}}

\textbf{Proposition \ref{prop:inefficiency_toy}. }[Inefficiency of LoRA fine-tuning]
\emph{Assume that LoRA weights are initialized with \init{1} or \init{2} and trained with gradient descent with learning rate $\eta = \Theta(n^c)$ for some $c \in \reals$. Then, it is impossible to have $\delta_t^i = \Theta(1)$ for all $i$ for any $t>0$, and therefore, fine-tuning with LoRA in this setup is inefficient.}

\begin{proof}
Assume that the model is initialized with \init{1}. Since the training dynamics are mainly simple linear algebra operation (matrix vector products, sum of vectors/scalars etc), it is easy to see that any vector/scaler in the training dynamics has a magnitude of order $n^{\gamma}$ for some $\gamma \in \reals$ (for more details, see the Tensor Programs framework, e.g. \cite{yang2019scaling}). For any quantity $v$ in the training dynamics, we write $v = \Theta(n^{\gamma[v]})$. When $v$ is a vector, we use the same notation when all entries of $v$ are $\Theta(n^{\gamma[v]})$. Efficiency is defined by having $\delta^t_i = \Theta(1)$ for $i \in \{1,2\}$ and $t>1$. Note that this implies $f_t(x) = \Theta(1)$ for all $t>1$. Let $t>1$ and assume that learning with LoRA is efficient. We will show that this leads to a contradiction. Efficiency requires that $\delta_t^i = \Theta(1)$ for all $t, i \in \{1,2\}$. Using the elementary formulas from \cref{sec:gamma_fct}, this implies that for all $t$
\begin{equation*}
    \begin{cases}
        \gamma[\eta]+2\gamma[b_{t-1}] + 1 = 0\\
        \gamma[\eta] + 2 \gamma[a_{t-1}^\top x] = 0\\
        \gamma[b_{t-1}] + \gamma[a_{t-1}^\top x] = 0.       
    \end{cases}
\end{equation*}
Solving this equation yields $\gamma[\eta] = -1/2$, i.e. LoRA finetuning can be efficient only if the learning rate scales as $\eta = \Theta(n^{-1/2})$. Let us now show that this yields a contradiction. From the gradient updates and the elementary operations from \cref{sec:gamma_fct}, we have the following recursive formulas

\begin{equation*}
    \begin{cases}
        \gamma[b_t] = \max(\gamma[b_{t-1}], -1/2 + \gamma[a_{t-1}^\top x]) \\
        \gamma[a_t^\top x] = \max(\gamma[a_{t-1}^\top x], 1/2+\gamma[b_{t-1}]) 
    \end{cases}
\end{equation*}
Starting from $t=1$, with Init[1] we have $\gamma[b_1] = \gamma[\eta (a_0^\top x)y] = -1/2$ and $\gamma[a_1^\top x] = \gamma[a_0^\top x] = 0$, we have $\gamma[b_2] = -1/2$ and $\gamma[a_2^\top x] = 0$. Trivially, this holds for any $t$. However, this implies that $\gamma[f_t] = \gamma[b_t] + \gamma[a_{t}^\top x] = -1/2$ which means that $\Delta f_t$ cannot be $\Theta(1)$. With Init[2], we have $\gamma[b_1] = \gamma[b_0] = 0$ and $\gamma[a_1^\top] = \gamma[\eta b_0 y \|x\|^2] = -1/2+1 = 1/2$. From the recursive formula we get $\gamma[b_2] = 0$ and $\gamma[a_2^\top x] = 1/2$ which remains true for all $t$. In this case we have $\gamma[f_t] = 1/2$ which contradicts $\Delta f_t = \Theta(1)$.

In both cases, this contradicts our assumption, and therefore efficiency cannot be achieved in this setup.
\\

\end{proof}

\subsection{Proof of \texorpdfstring{\cref{prop:efficient_toy}}{Proposition Efficient Toy}}

\textbf{Proposition \ref{prop:efficient_toy}.} [Efficient Fine-Tuning with LoRA]
\emph{In the case of Toy model \cref{eq:toy_model}, with $\eta_a = \Theta(n^{-1})$ and $\eta_b = \Theta(1)$, we have for all $t>1$, $\in \{1,2,3\}$, $\delta_t^i = \Theta(1)$.}

\begin{proof}
The proof is similar in flavor to that of \cref{prop:inefficiency_toy}. In this case, the set of equations that should be satisfied so that $\delta_t^i = \Theta(1)$ are given by
\begin{equation*}
    \begin{cases}
        \gamma[\eta_a]+2\gamma[b_{t-1}] + 1 = 0\\
        \gamma[\eta_b] + 2 \gamma[a_{t-1}^\top x] = 0\\
        \gamma[\eta_a] + \gamma[\eta_b]+\gamma[b_{t-1}] + \gamma[a_{t-1}^\top x]+1 = 0,        
    \end{cases}
\end{equation*}
where we have used the elementary formulas from \cref{sec:gamma_fct}. Simple calculations yield $\gamma[\eta_a] + \gamma[\eta_b] = -1$. Using the gradient update expression with the elementary addition from \cref{sec:gamma_fct}, the recursive formulas controlling $\gamma[b_t]$ and $\gamma[a_t^\top x]$ are given by
\begin{equation*}
    \begin{cases}
        \gamma[b_t] = \max(\gamma[b_{t-1}], \gamma[\eta_b] + \gamma[a_{t-1}^\top x]) \\
        \gamma[a_t^\top x] = \max(\gamma[a_{t-1}^\top x], \gamma[\eta_a]+\gamma[b_{t-1}]+1). 
    \end{cases}
\end{equation*}

Starting from $t=1$, with \init{1}, we have $\gamma[b_1] = \gamma[\eta_b (a_0^\top x) y] = \gamma[\eta_b]$ and $\gamma[a_1^\top x] = \gamma[a_0^\top x] = 0$. Therefore $\gamma[b_2] = \max(\gamma[\eta_b], \gamma[\eta_b] + 0) = \gamma[\eta_b]$, and $\gamma[a_2^\top x] = \max(0, \gamma[\eta_a] + \gamma[\eta_b] +1) = \max(0,0) = 0$. By induction, this holds for all $t\geq 1$. With \init{2}, we have $\gamma[b_1] = \gamma[b_0] = 0$, and $\gamma[a_1^\top x] = \gamma[-\eta_a b_0^2 y \|x\|^2] = \gamma[\eta_a] + 1$. At step $t=2$, we have $\gamma[b_2] = \max(0, \gamma[\eta_b] + \gamma[\eta_a] + 1) = 0$ and $\gamma[a_2^\top x] = \max(\gamma[\eta_a] + 1, \gamma[\eta_a] + 0 + 1) = \gamma[\eta_a] + 1$, and this holds for all $t$ by induction. In both cases, to ensure that $\gamma[f_t] = \gamma[b_t] + \gamma[a_t^\top x] = 0$, we have to set $\gamma[\eta_b] = 0$ and $\gamma[\eta_a]=-1$ (straightforward from the equation $\gamma[\eta_b] + \gamma[\eta_a] = -1$). In conclusion, setting $\eta_a = \Theta(n^{-1})$ and $\eta_b = \Theta(1)$ ensures efficient fine-tuning with LoRA.

\end{proof}

\subsection{Proof of \texorpdfstring{\cref{thm:efficiency}}{Theorem Efficiency}}

In this section, we give a non-rigorous but intuitive proof of \cref{thm:efficiency}. The proof relies on the following assumption on the processed gradient $g_A$.

\begin{assumption}\label{assumption}
With the same setup of \cref{sec:main_theory}, at training step $t$, we have $g_A^{t} \underbar{Z} = \Theta(n)$.    
\end{assumption}
To see why \cref{assumption} is sound in practice, let us study the product $g_A^{t} \underbar{Z}$ in the simple case of Adam with no momentum, a.k.a SignSGD which is given by  

$$
g_A = \textrm{sign}\left(\frac{\partial \loss}{\partial A}\right),
$$
where the sign function is applied element-wise. At training step $t$, we have 
$$
\frac{\partial \loss_{t}}{\partial A} = \frac{\alpha}{r} B^\top_{t-1} d\Bar{Z}^{t-1} \otimes \underbar{Z},
$$
Let $S^t = \frac{\alpha}{r} B^\top_{t-1} d\Bar{Z}^{t-1}$. Therefore we have 
$$
g_A = \textrm{sign}(S^t \otimes \underbar{Z}) = (\textrm{sign}(S^t_{i} \underbar{Z}_j))_{1\leq i, j \leq n}.
$$

However, note that we also have 
$$
\textrm{sign}(S^t_{i} \underbar{Z}_j) = \textrm{sign}(S^t_{i}) \textrm{sign}(\underbar{Z}_j),
$$
and as a result  
$$
g_A^{t} = \textrm{sign}(S^t) \otimes \textrm{sign}(\underbar{Z}).
$$

Hence, we obtain 
$$
g_A^{t} \underbar{Z} = (\textrm{sign}(\underbar{Z})^\top \underbar{Z} ) \textrm{sign}(S^t) = \Theta(n),
$$
where we used the fact that $\textrm{sign}(\underbar{Z})^\top \underbar{Z} = \Theta(n)$.\\

This intuition should in-principle hold for the general variant of Adam with momentum as long as the gradient processing function (a notion introduced in \cite{yang2013theory}) roughly preserves the $\textrm{sign}(\underbar{Z})$ direction. This reasoning can be made rigorous for general gradient processing function using the Tensor Program framework and taking the infinite-width limit where the components of $g_A, \underbar{Z}, d\bar{Z}$ all become iid. However this necessitates an intricate treatment of several quantities in the process, which we believe is an unnecessary complication and does not serve the main purpose of this paper. 

Let us now give a proof for the main claim. 

\textbf{Theorem \ref{thm:efficiency}.}
\emph{Assume that weight matrices $A$ and $B$ are trained with Adam with respective learning rates $\eta_A$ and $\eta_B$ and that \cref{assumption} is satisifed with the Adam gradient processing function. Then, it is impossible to achieve efficiency with $\eta_A = \eta_B$. However, LoRA Finetuning is efficient with $\eta_A = \Theta(n^{-1})$ and $\eta_B = \Theta(1)$.}\\

\begin{proof}

With the same setup of \cref{sec:main_theory}, at step $t$, we have 
\begin{equation*}
    \begin{cases}
        \delta_t^1 = B_{t-1} \Delta Z_A^t  = -\eta_A B_{t-1} g^{t-1}_A \underbar{Z}\\
        \delta_t^2 = \Delta B_t Z_A^{t-1} = -\eta_B g^{t-1}_B A_{t-1} \underbar{Z}\\
        \delta_t^3 = \Delta B_t \Delta Z_A^t = \eta_A \eta_B g^{t-1}_B g^{t-1}_A \underbar{Z}
    \end{cases}
\end{equation*}

The key observation here is that $g^{t-1}_A \underbar{Z}$ has entries of order $\Theta(n)$ as predicted and justified in \cref{assumption}. Having $\delta^i_t = \Theta(1)$ for $i \in \{1,2\}$ and $Z_B^t = \Theta(1)$ for $t >1$ translate to 
\begin{equation*}
\begin{cases}
    \gamma[\eta_A] + \gamma[B_{t-1}]+1  = 0\\
    \gamma[\eta_B] + \gamma[A_{t-1} \underbar{Z}] = 0\\
    \gamma[B_{t-1}] + \gamma[A_{t-1} \underbar{Z}] = 0,
\end{cases}
\end{equation*}
which implies that $\gamma[\eta_A] + \gamma[\eta_B] = -1$.

With the gradient updates, we have 
\begin{align*}
    B_t &= B_{t-1} - \eta_B g^{t-1}_B\\
    A_t \underbar{Z} &= A_{t-1}\underbar{Z} - \eta_A g^{t-1}_A \underbar{Z}
\end{align*}
which implies that 
\begin{align*}
    \gamma[B_t] &= \max(\gamma[B_{t-1}], \gamma[\eta_B])\\
    \gamma[A_t \underbar{Z}] &= \max(\gamma[A_{t-1} \underbar{Z}], \gamma[\eta_A]+1),
\end{align*}

Now assume that the model is initialized with \init{1}. We have $\gamma[B_1] = \gamma[\eta_B]$ and therefore for all $t$, we have $\gamma[B_t] = \gamma[\eta_B]$. We also have $\gamma[A_1 \underbar{Z}] = \gamma[A_0 \underbar{Z}] = 0$ (because $A_1 = A_0$, and we use the Central Limit Theorem to conclude). Hence, if we choose the same learning rate for $A$ and $B$, given by $\eta$, we obtain $\gamma[\eta] = -1/2$, and therefore $\gamma[Z_A^{t-1}] = \gamma[A_{t-1} \underbar{Z}] = 1/2$ which violates the stability condition. A similar behaviour occurs with \init{2}. Hence, efficiency is not possible in this case. However, if we set $\gamma[\eta_B] = 0$ and $\gamma[\eta_A] = -1$, we get that $\gamma[B_t] = 0, \gamma[A_t \underbar{Z}] = 0$, and $\delta_t^i = \Theta(1)$ for all $i\in \{1,2,3\}$ and $t\geq 1$. The same result holds with \init{2}. 

\end{proof}

\section{Efficiency from a Loss Perspective.}
Consider the same setup of \cref{sec:main_theory}. At step $t$, the loss changes as follows
\begin{align*}
\Delta \loss &= \loss((BA)_t) - \loss((BA)_{t-1}) \\
&\approx \langle d \Bar{Z}^{t-1} \otimes \underbar{Z}, (BA)_t - (BA)_{t-1} \rangle_F \\
&= \langle d\Bar{Z}^{t-1}, \Delta Z_B^t \rangle,
\end{align*}    
where $\langle.,.\rangle_F$ is the Frobenius inner product in $\reals^{n\times n}$, and $\langle.,.\rangle$ is the euclidean product in $\reals^n$. Since the direction of the feature updates are significantly correlated with $d \Bar{Z}^{t-1}$, it should be expected that having $\delta_t^i = \Theta(1)$ for all $i$ results in more efficient loss reduction.

\section{Additional Experiments}\label{app:add_exps}
This section complements the empirical results reported in the main text. We provide the details of our experimental setup, and show the acc/loss heatmaps for several configurations.
\subsection{Empirical Details}
\subsubsection{Toy Example}
In \cref{fig:toy_heatmap}, we trained a simple MLP with LoRA layers to verify the results of the analysis in \cref{sec:toy}. Here we provide the empirical details for these experiments.

\paragraph{Model.} We consider a simple MLP given by
\begin{equation*}
f(x) = W_{out} \phi(BA \phi(W_{in} x)),
\end{equation*}
where $W_{in} \in \reals^{n\times d}, W_{out} \in \reals^{1\times n}, A \in \reals^{r\times n}, B \in \reals^{n \times r}$ are the weights, and $\phi$ is the ReLU activation function. Here, we used $d=5$, $n=100$, and $r=4$.

\paragraph{Dataset.} Synthetic dataset generated by $X\sim \normal(0, I_d), Y = \sin(d^{-1}\sum_{i=1}^d X_i)$ with $d=5$. The number of training examples is $N_{train}=1000$, and the number of test examples is $N_{test}=100$.

\paragraph{Training.} We train the model with gradient descent for a range for values of $(\eta_A,\eta_B)$. The weights are initialized as follows: $W_{in} \sim \normal(0,1.), W_{out} \sim \normal(0, 1/n), A \sim \normal(0, 1/n), B \sim \normal(0, 1.)$. Only the weight matrices $A, B$ are trained and $W_{in}, W_{out}$ are fixed to their initial value.

\subsubsection{GLUE Tasks with GPT2/Roberta}

For our experiments with GPT2/Roberta-base models, finetuned on GLUE tasks, we use the following setup:

\paragraph{Tasks. } MNLI, QQP, SST2, QNLI

\paragraph{Models.} GPT2, Roberta-base

\paragraph{Training Alg.} AdamW with $\beta_1=0.9, \beta_2=0.99, \epsilon = $ 1e-8, linear schedule, no warmup. 

\paragraph{Learning rate grid.} $\eta_A \in \{$4e-3, 2e-3, 1e-3, 5e-4, 2e-4, 1e-4$\}$, $\eta_B \in \{$ 8e-4, 4e-4, 2e-4, 1e-4, 5e-5, 2e-5, 1e-5 $\}$.

\paragraph{Targert Modules for LoRA.} For Roberta-base, we add LoRA layers to `query' and `value' weights. For GPT2, we add LoRA layers to `c\_attn, c\_proj, c\_fc'.

\paragraph{Other Hyperparameters.} Sequence length $T=128$, train batch size $bs = 32$, number of train epochs $E=3$ ($E=10$ for SST2), number of random seeds $s=3$.

\paragraph{GPUs. } Nvidia V100, Nvidia A10.

\subsubsection{Llama MNLI}
For our experiments using the Llama-7b model, finetuned on MNLI, we use following setup

\paragraph{Training Alg.} AdamW with $\beta_1 = 0.9$, $\beta_2 = 0.999$, $\epsilon =$ 1e-6, constant schedule.

\paragraph{Learning rate grid.} $\eta_A \in \{$1e-6, 5e-6, 1e-5, 2.5e-5, 5e-5, 1e-4$\}$, $\eta_B \in \{$1e-6, 5e-6, 1e-5, 2.5e-5, 5e-5, 1e-4$\}$, $\eta_B \geq \eta_A$

\paragraph{LoRA Hyperparameters.} LoRA rank $r = 8$, $\alpha = 16$, and dropout $0.1$. LoRA target modules `q\_proj, k\_proj, v\_proj, o\_proj, up\_proj, down\_proj, gate\_proj'.

\paragraph{Other Hyperparameters.} Sequence length $T = 128$, train batch size $bs = 32$, number of train epochs $E = 1$, number of random seeds $s = 2$ for $\eta_A = \eta_B$ and $\eta_A, \eta_B$ near test optimal, $s = 1$ otherwise. Precision FP16. 

\paragraph{GPUs.} Nvidia V100.

\subsubsection{Llama flan-v2}

For our experiments using the Llama-7b model, finetuned on a size 100k random subset flan-v2, we use following setup

\paragraph{Training Alg.} AdamW with $\beta_1 = 0.9$, $\beta_2 = 0.999$, $\epsilon =$ 1e-6, constant schedule.

\paragraph{Learning rate grid.} $\eta_A \in \{$1e-6, 5e-6, 1e-5, 2.5e-5, 5e-5, 1e-4$\}$, $\eta_B \in \{$1e-6, 5e-6, 1e-5, 2.5e-5, 5e-5, 1e-4$\}$, $\eta_B \geq \eta_A$

\paragraph{LoRA Hyperparameters.} LoRA rank $r = 64$, $\alpha = 16$, and dropout $0.1$. LoRA target modules `q\_proj, k\_proj, v\_proj, o\_proj, up\_proj, down\_proj, gate\_proj'.

\paragraph{Other Hyperparameters.} Sequence length $T_{\text{source}} = 1536$, $T_{\text{target}} = 512$, train batch size $bs = 16$, number of epochs $E = 1$, number of random seeds $s = 2$ for $\eta_A = \eta_B$ and $\eta_A, \eta_B$ near test optimal, $s = 1$ otherwise. Precision BF16.

\paragraph{MMLU Evaluation.} We evaluate average accuracy on MMLU using 5-shot prompting.

\paragraph{GPUs.} Nvidia A10.

\subsection{Results of Roberta-base Finetuning on all Tasks}
\cref{fig:roberta_main} showed finetuning test accuracy for Roberta-base. To complement these results, we show here the test/train accuracy for all tasks. 

\begin{figure}[h]
    \centering
    \includegraphics[width=0.95\linewidth]{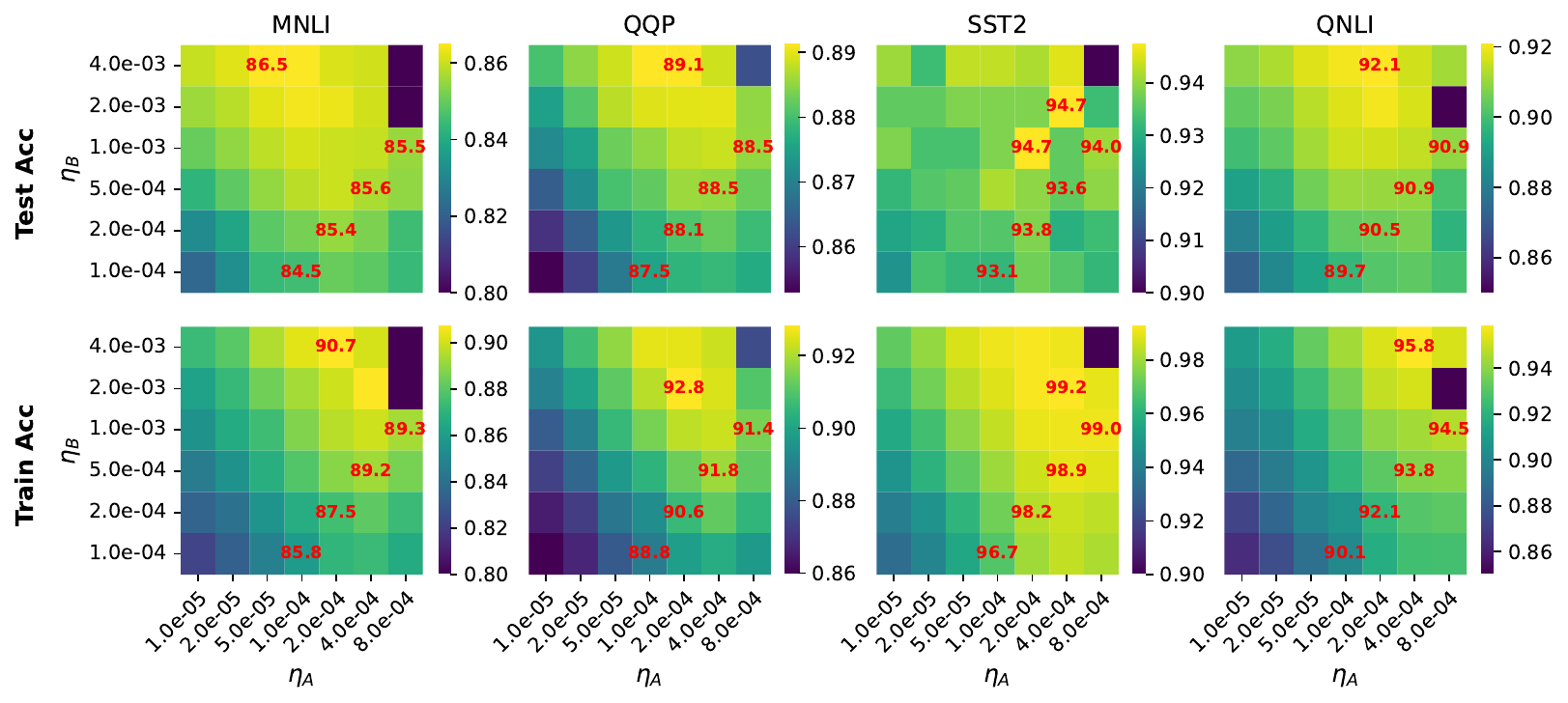}
    \caption{GLUE/Roberta-base: same as \cref{fig:roberta_main} with test/train accuracy.}
    \label{fig:enter-label}
\end{figure}

Interestingly, the optimal choice of learning rates for test accuracy differs from that of the train accuracy, although the difference is small. This can be due to mild overfitting occuring during finetuning (the optimal choice of learning rates $(\eta_A, \eta_B)$ for train accuracy probably lead to a some overfitting). 

\subsection{Results of GPT2 Finetuning on all Tasks}
\cref{fig:gpt2_main} showed finetuning results for GPT2 on MNLI and QQP. To complement these results, we show here the test/train accuracy for all tasks. 

\begin{figure}[h]
    \centering
    \includegraphics[width=0.9\linewidth]{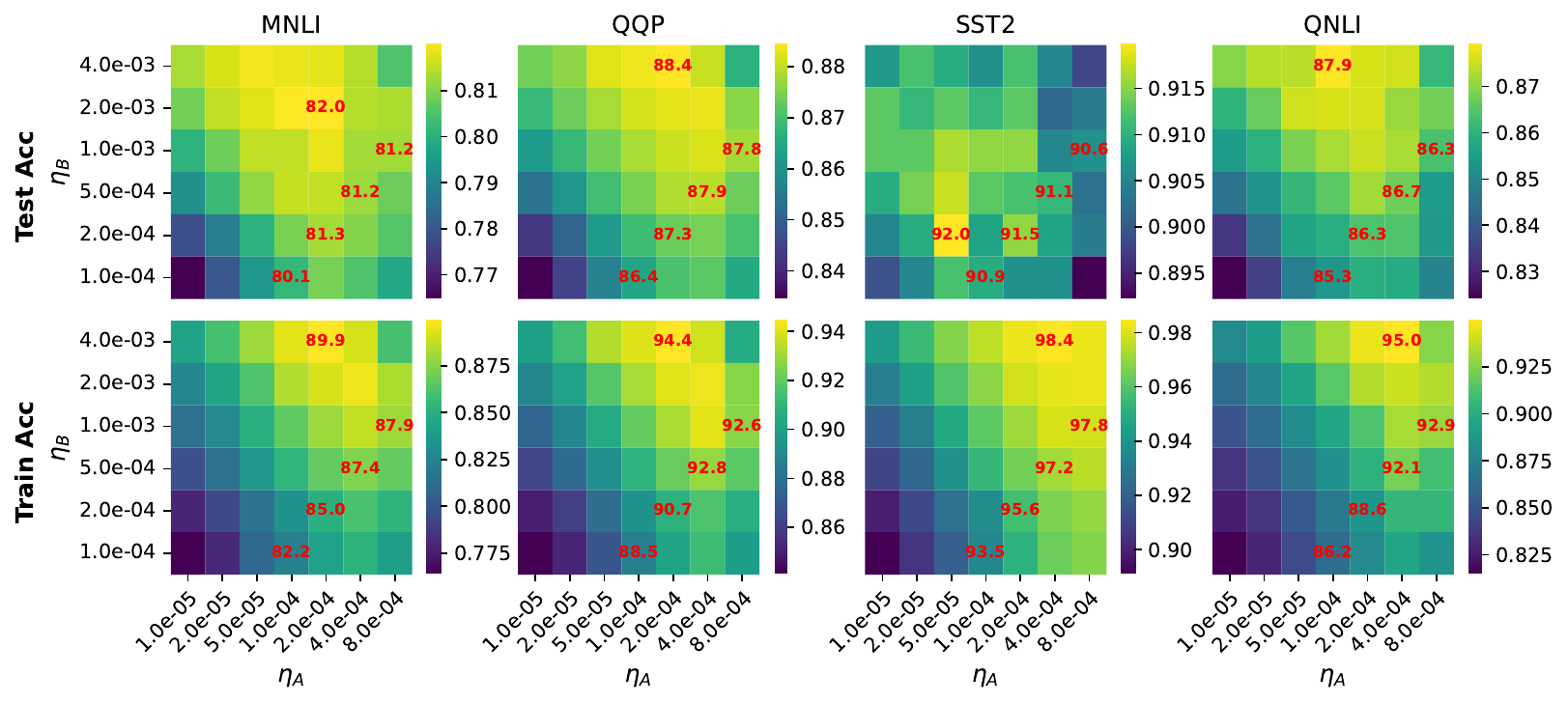}
    \caption{GLUE/GPT2: same setup as \cref{fig:gpt2_main} with additional tasks}
    \label{fig:gpt2_all_tasks_half_precision}
\end{figure}

\newpage
\subsection{GLUE Tasks with Full Precision}

\begin{figure}[h]
    \centering
 \includegraphics[width=0.9\linewidth]{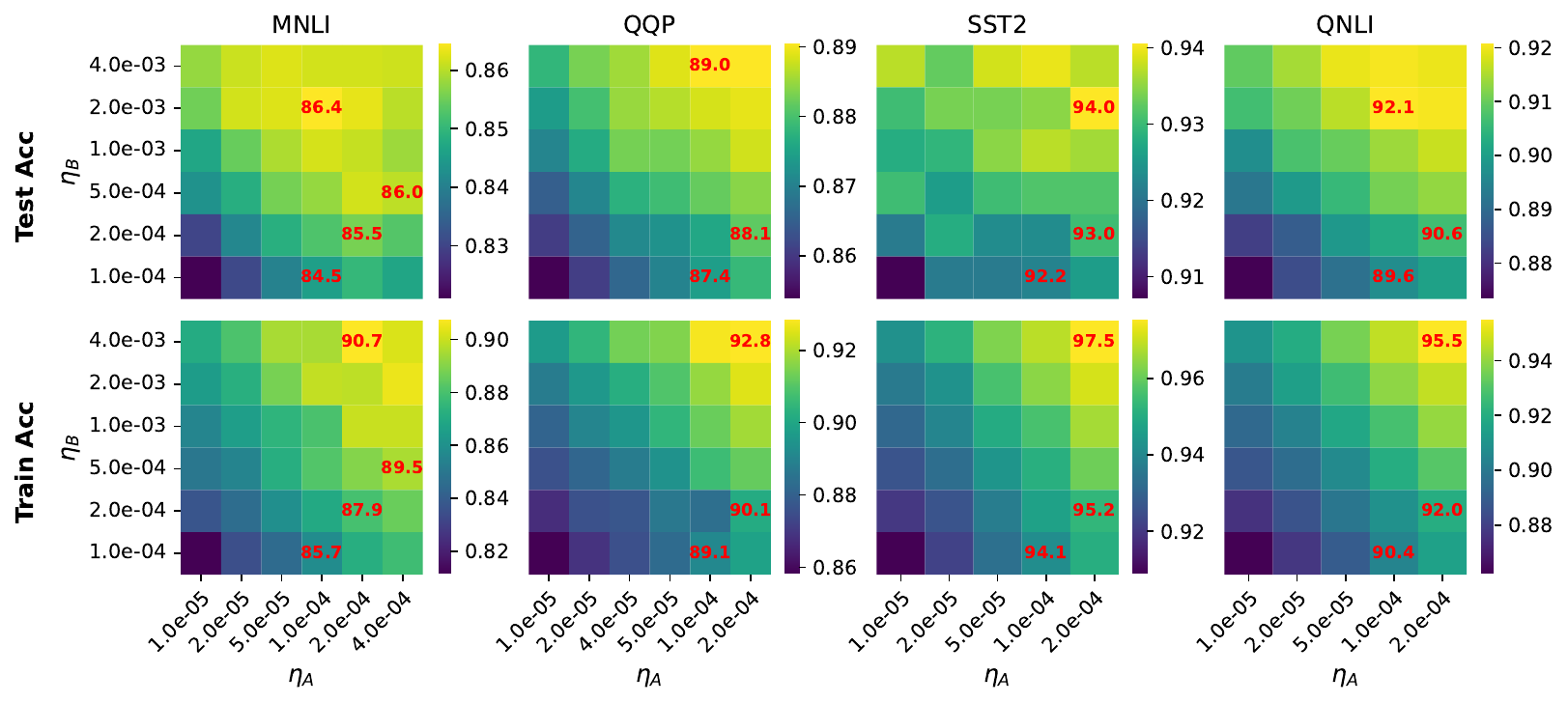}
    \caption{GLUE/Roberta-base: same as \cref{fig:roberta_main} with full precision training instead of FP16.}
    \label{fig:full_precision_glue_roberta}
\end{figure}

\begin{figure}[h]
    \centering
    \includegraphics[width=0.9\linewidth]{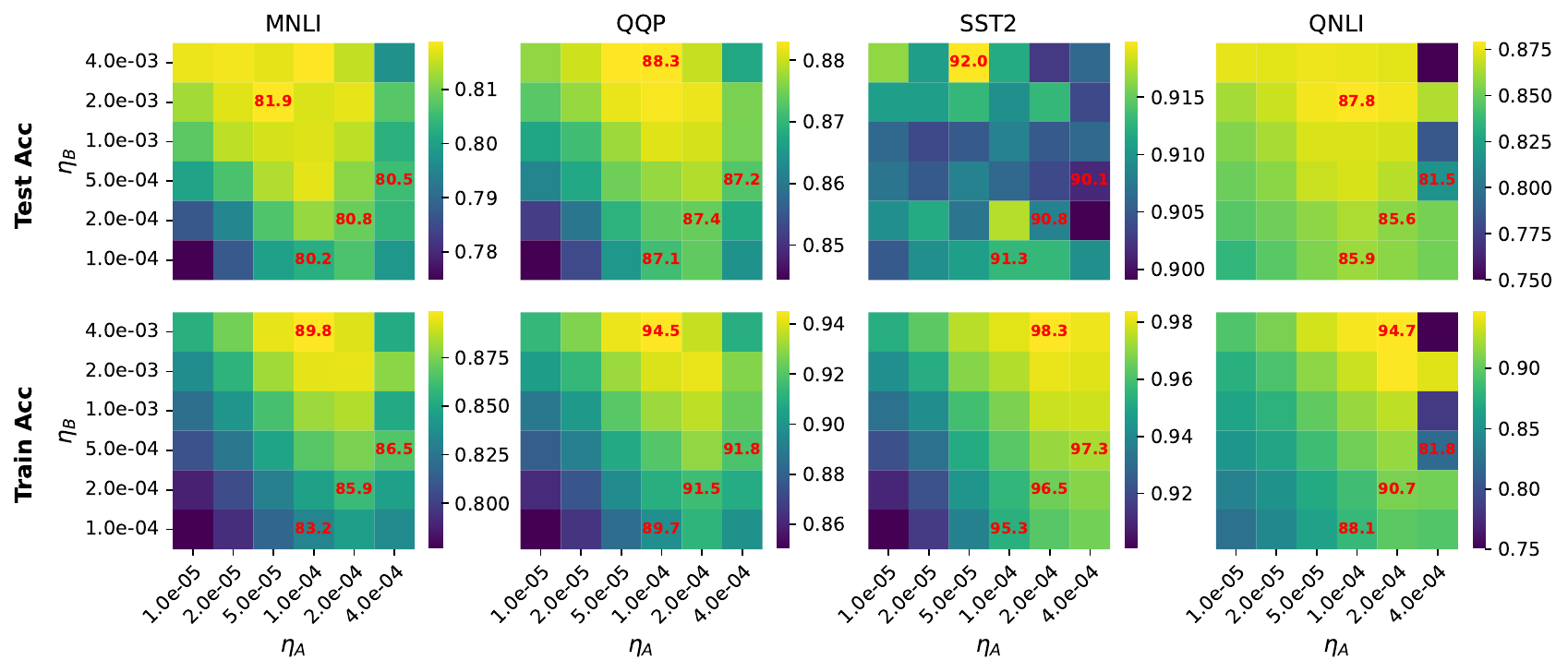}
    \caption{GLUE/GPT2: same setup as \cref{fig:gpt2_all_tasks_half_precision} with full precision training}
    \label{fig:gpt2_all_tasks_full_precision}
\end{figure}

\newpage
\subsection{GLUE Tasks Test/Train Loss}

\begin{figure}[h]
    \centering
    \includegraphics[width=0.9\linewidth]{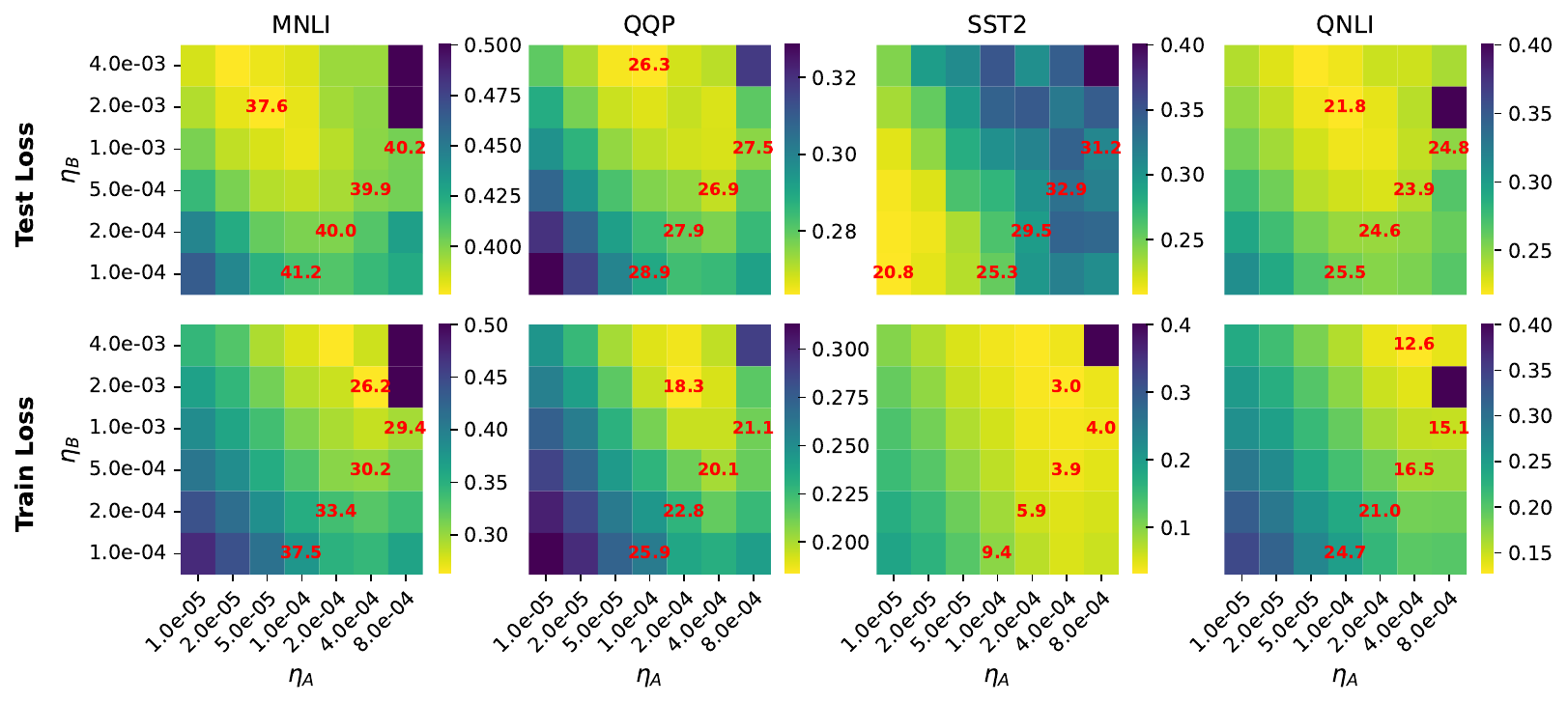}
    \caption{GLUE/Roberta-base: same setup as \cref{fig:roberta_main} with $100\times$Test/Train loss instead of accuracy}
    \label{fig:full_precision_glue_roberta_loss}
\end{figure}

\begin{figure}[h]
    \centering
    \includegraphics[width=0.9\linewidth]{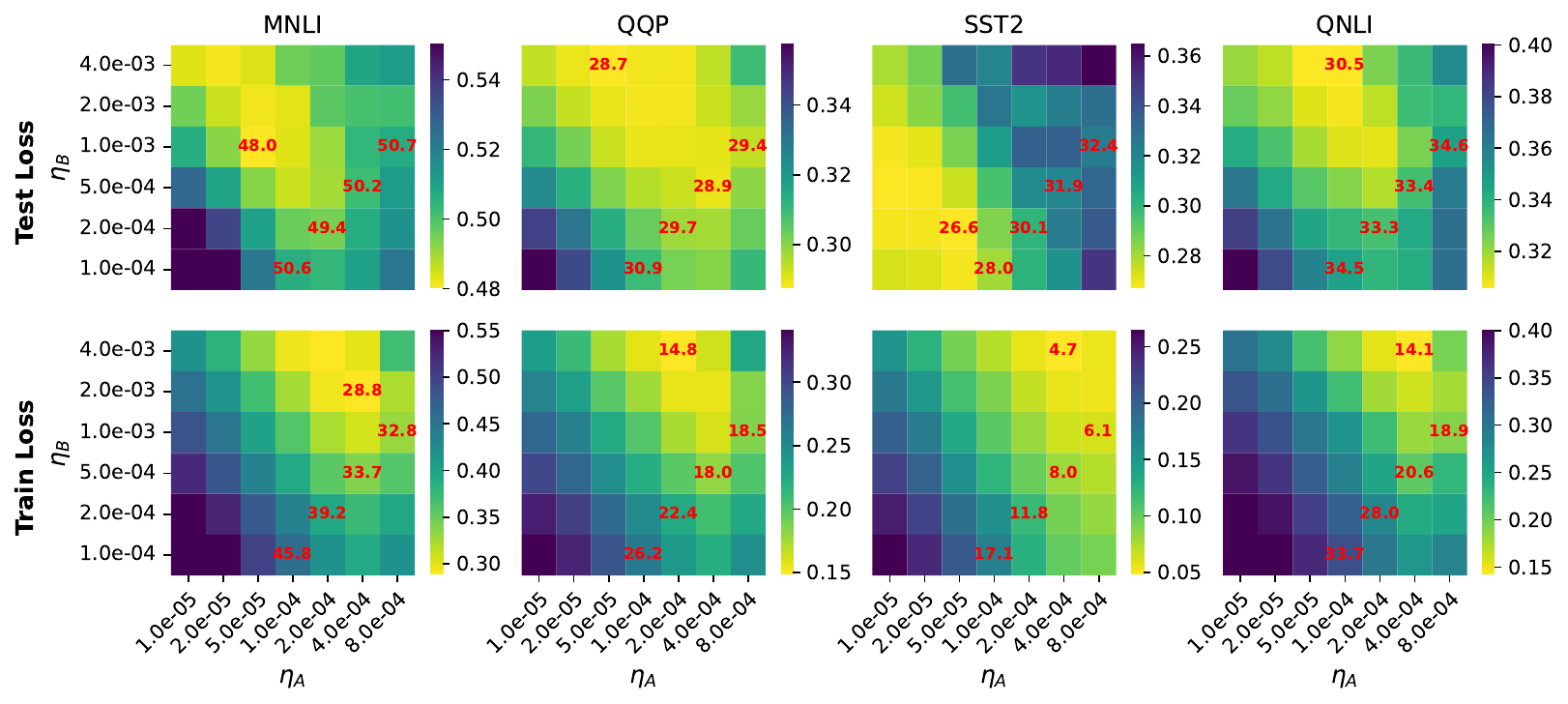}
    \caption{GLUE/GPT2: same setup as \cref{fig:gpt2_all_tasks_half_precision} with $100\times$Test/Train loss instead of accuracy}
    \label{fig:gpt2_halprecision_alltasks_loss}
\end{figure}

\newpage
\subsection{GLUE Tasks with Different LoRA Ranks}

\vspace{-2mm}

\begin{figure}[h!]
    \centering
    \includegraphics[width=0.7\linewidth]{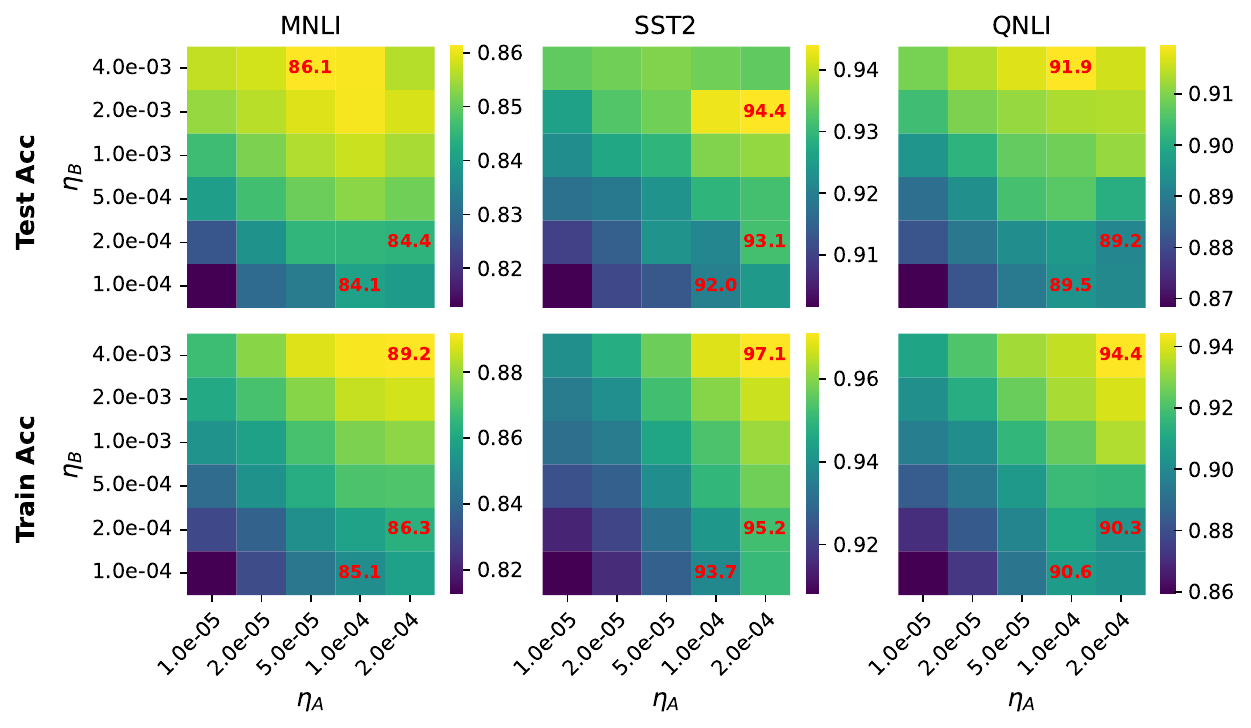}
    \vspace{-2mm}
    \caption{GLUE/Roberta-base: same setup as \cref{fig:roberta_main} with $r=4$}
    \label{fig:roberta_rank4}
\end{figure}

\vspace{-4mm}
\begin{figure}[h!]
    \centering
    \includegraphics[width=0.45\linewidth]{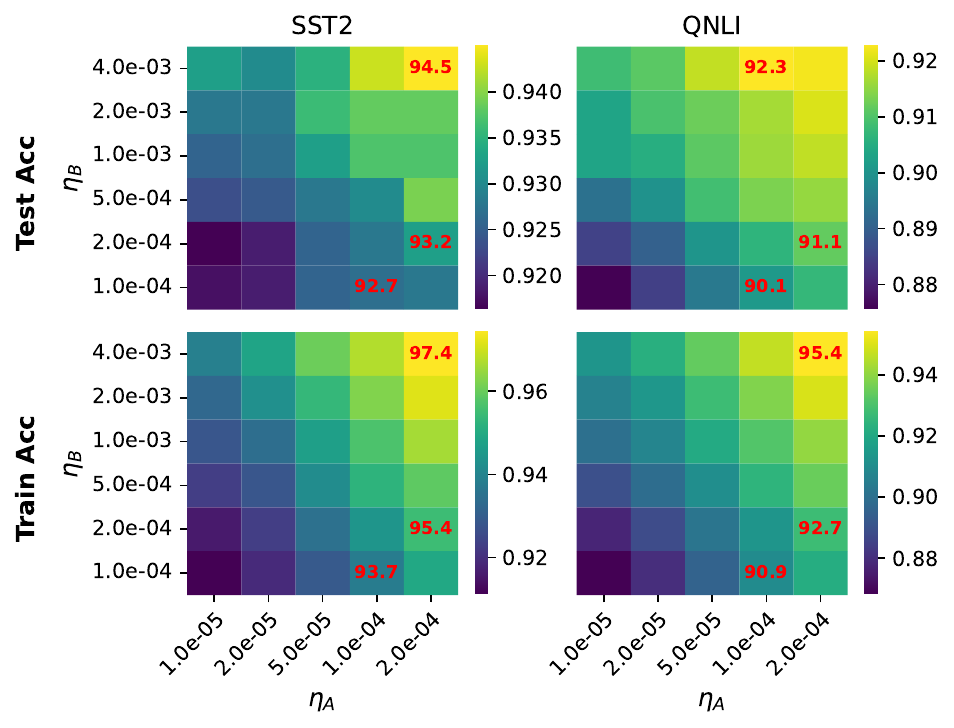}
    \vspace{-3mm}
    \caption{GLUE/Roberta-base: same setup as \cref{fig:roberta_main} with $r=16$}
    \label{fig:roberta_rank16}
\end{figure}

\vspace{-2mm}
\begin{figure}[h!]
    \centering
    \includegraphics[width=0.45\linewidth]{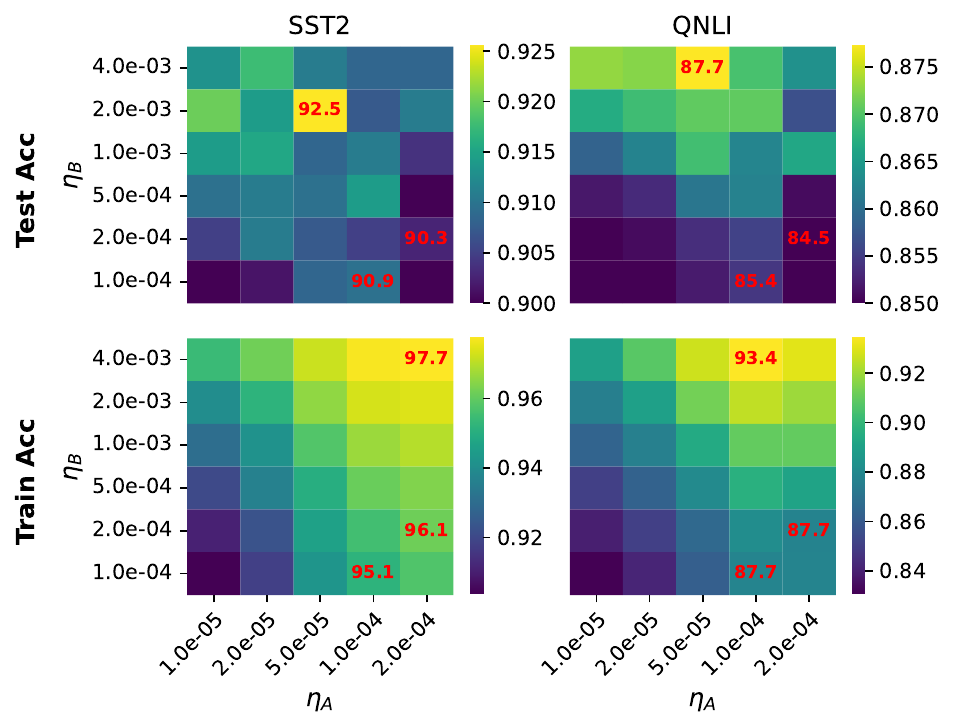}
    \vspace{-2mm}
    \caption{GLUE/GPT2: same setup as \cref{fig:gpt2_all_tasks_full_precision} with $r=4$}
    \label{fig:glue_rank4}
\end{figure}

\subsection{Experiments with \init{1}}
\begin{figure}
    \centering
    \includegraphics[width=0.35\linewidth]{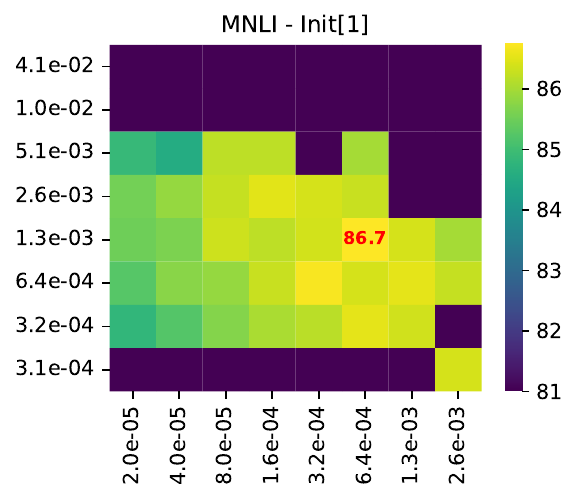}
    \includegraphics[width=0.35\linewidth]{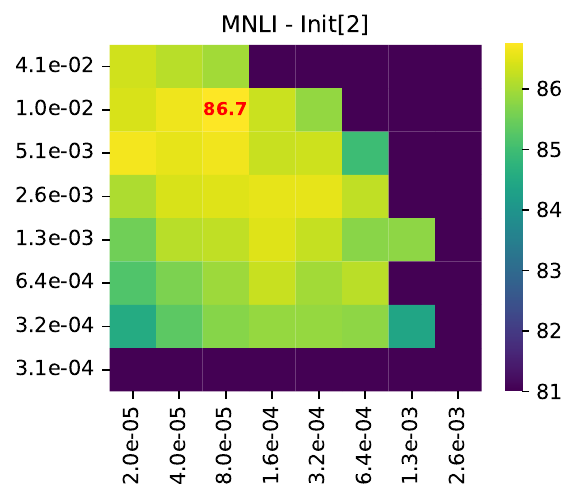}
    \caption{Roberta-base with \init{1} and \init{2}, finetuning on MNLI for 10 epochs (similar to \cref{fig:roberta_main} but with more epochs).}
    \label{fig:roberta-init1}
\end{figure}

We also run some experiments using \init{1} as initialization scheme. We noticed that the optimal ratio $\lambda$ is this case is generally smaller than the optimal ratio with \init{2}. \cref{fig:roberta-init1} shows the optimal learning rates $(\eta_A,\eta_B)$ obtained with \init{1} and \init{2}. The optimal ratio $\lambda=\eta_B/\eta_A$ is generally smaller with \init{1}. 

\newpage
\subsection{Llama Flan-v2 MMLU Acc/Train Loss}
\begin{figure}[!htb]
    \centering
    \begin{subfigure}{0.7\linewidth}
        \includegraphics[width=0.9\linewidth]{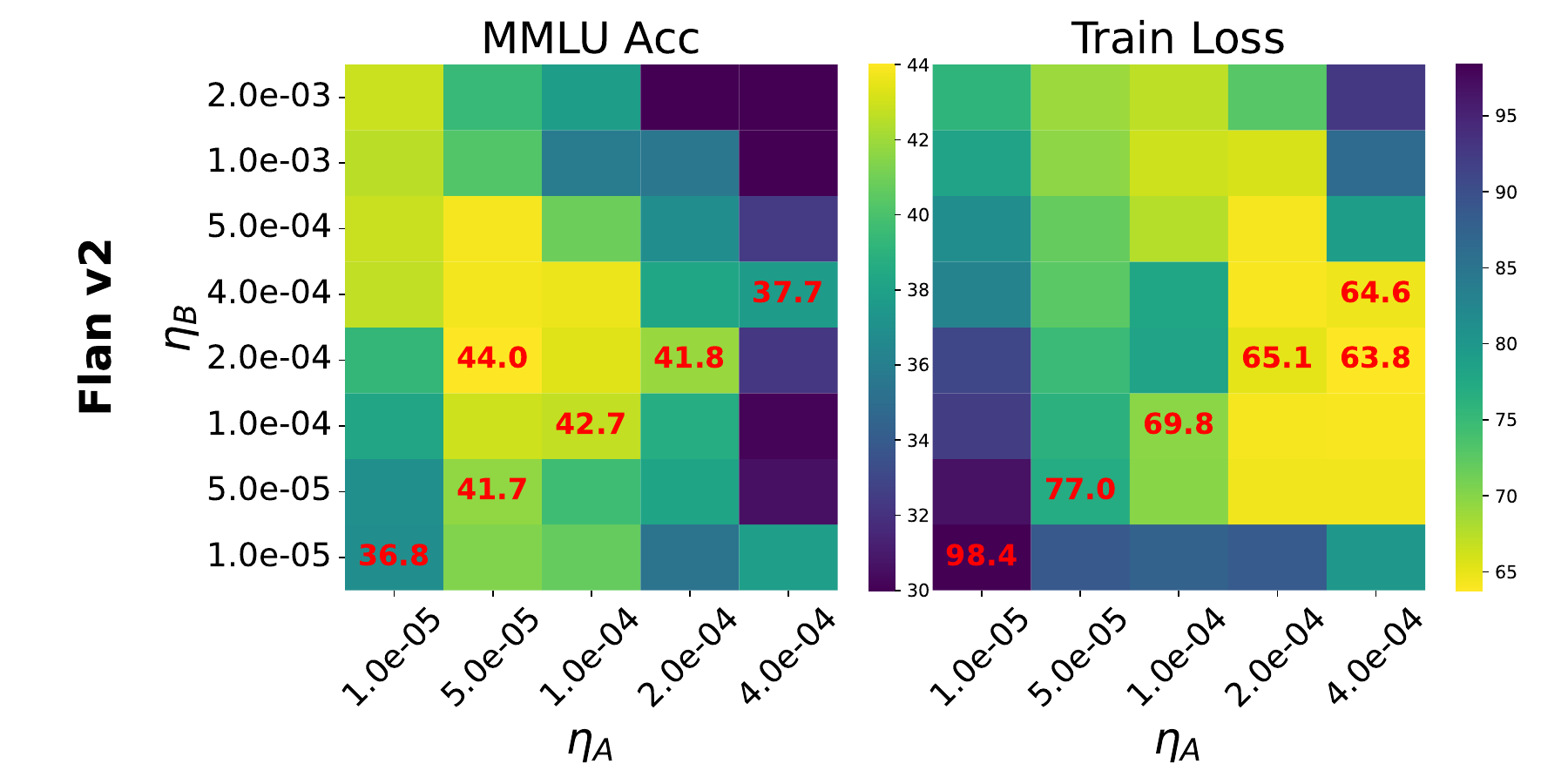}
        \caption{MMLU evaluation accuracy and train loss of Llama-7b trained on flan-v2 100k in the same setting as Figure \ref{fig:llama_test} left panel (using \init{2}). Interestingly, even in one epoch the model can overfit. We were unable to find $\eta_B > \eta_A$ that was optimal for train loss, however it could be the case that the grid was not fine enough or that overfitting does not require much ``feature learning" and $\eta_B / \eta_A \approx 1$ is optimal for minimizing train loss (see the main text for more discussion).\\}
        \label{fig:llama_flan_v2}
    \end{subfigure}
    \begin{subfigure}{0.7\linewidth}
        \includegraphics[width=0.9\linewidth]{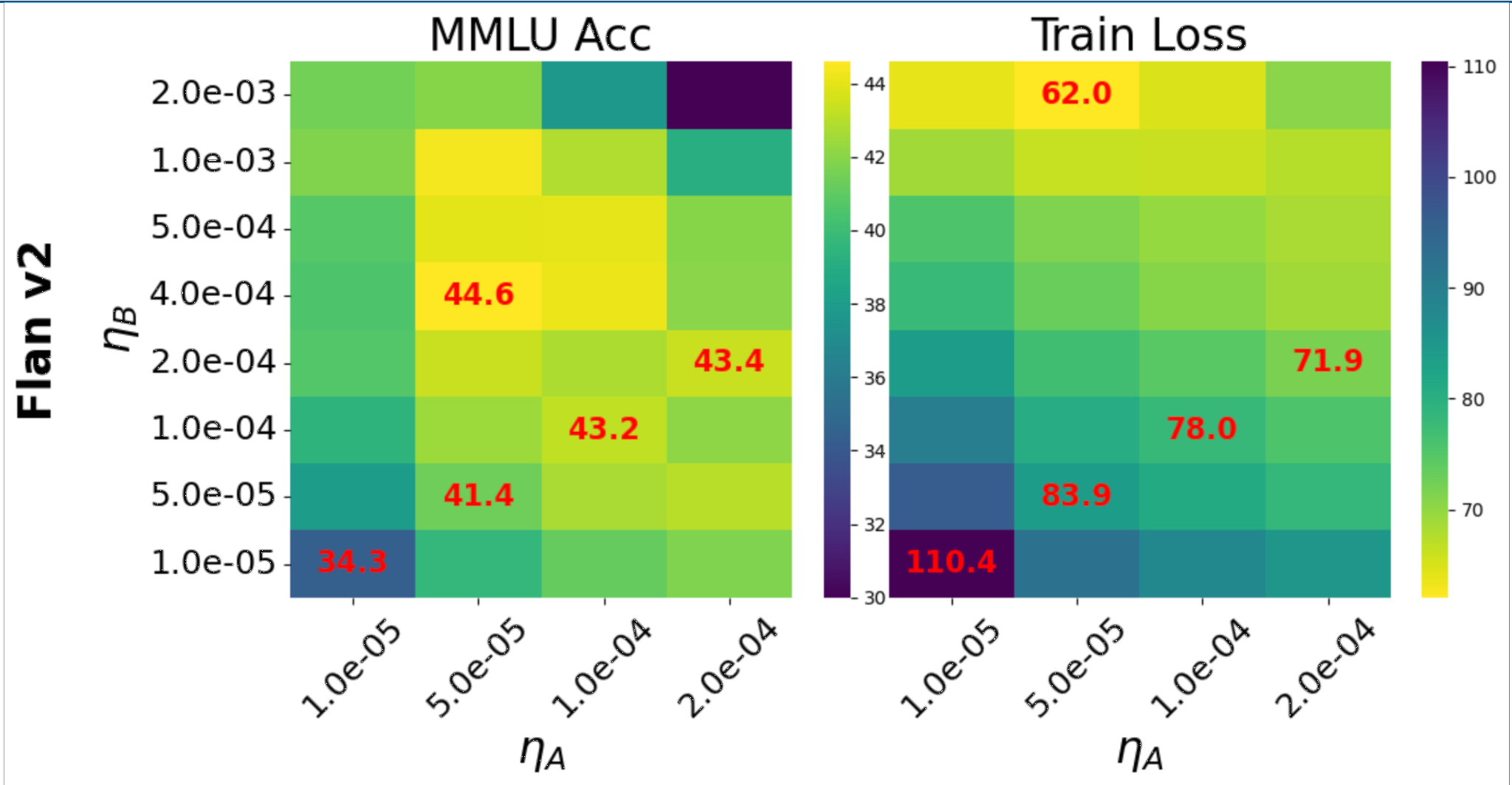}
        \caption{MMLU evaluation accuracy and train loss of Llama-7b trained on flan-v2 100k in the same setting as Figure \ref{fig:llama_test} left panel except using \init{1}. Interestingly, the optimal MMLU accuracy is 0.6\% higher than using \init{2} and the optimal ratio $\eta_B / \eta_A$ is twice as large. The training loss is also near optimal only using a large ratio $\eta_B / \eta_A$.}
        \label{fig:llama_flan_v2_init1}
    \end{subfigure}
    \caption{Llama-7b on flan-v2 training with different initializations.}
\end{figure}
\newpage 
\subsection{Llama MNLI Test/Train Loss}

\begin{figure}[h]
    \centering
    \includegraphics[width=0.7\linewidth]{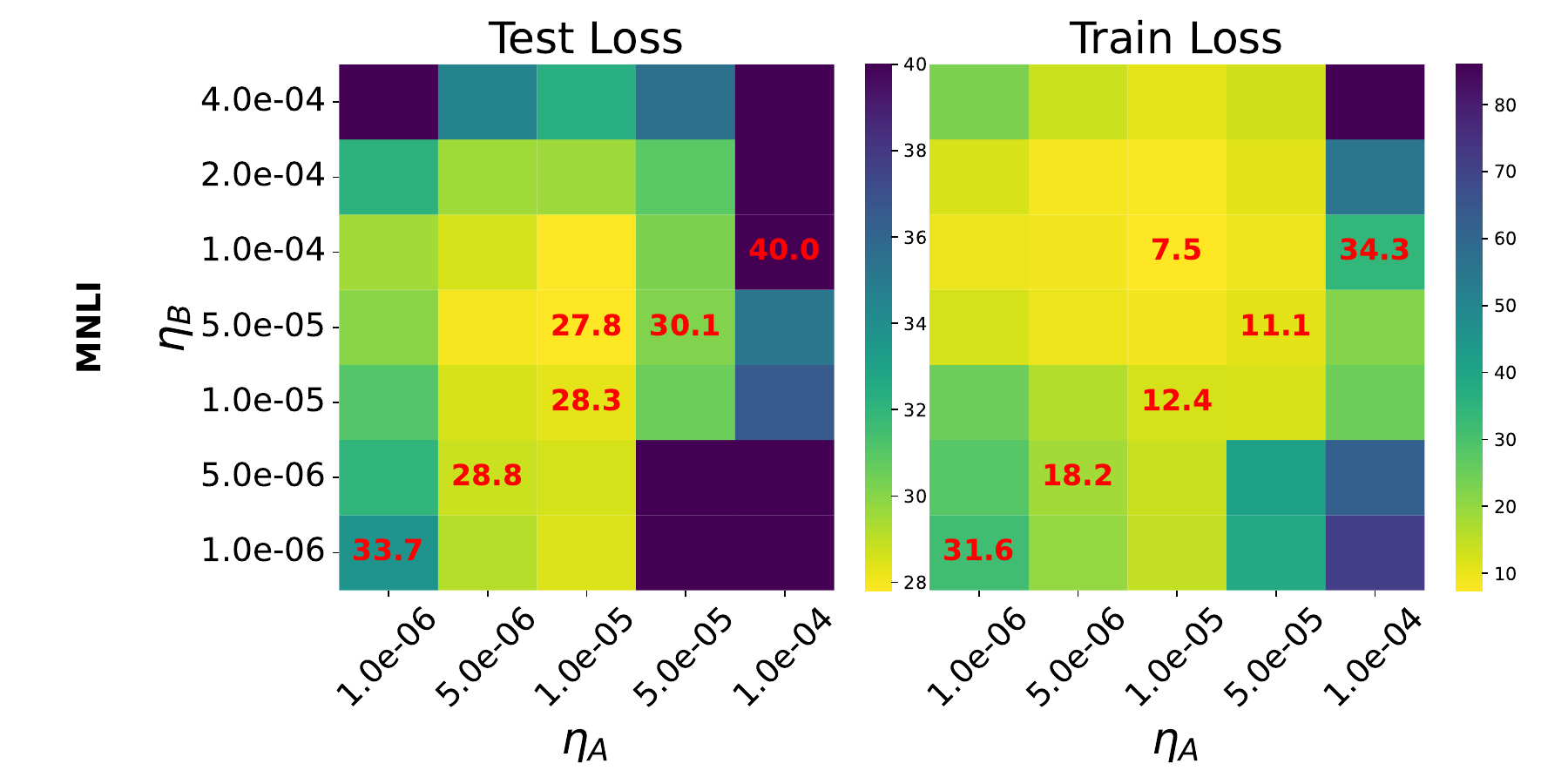}
    \caption{Train and test loss of Llama-7b finetuned on MNLI in the same setting as Figure \ref{fig:llama_test} right panel.}
    \label{fig:llama_mnli_loss}
\end{figure}


\end{document}